%% file: arxiv.tex
\documentclass[letterpaper]{article} 
\usepackage{aaai24}  
\usepackage{times}  
\usepackage{helvet}  
\usepackage{courier}  
\usepackage[hyphens]{url}  
\usepackage{graphicx} 
\usepackage{natbib}  
\usepackage{caption} 
\usepackage{algorithm}
\usepackage{algorithmic}

\usepackage{newfloat}
\usepackage{listings}
\DeclareCaptionStyle{ruled}{labelfont=normalfont,labelsep=colon,strut=off} 
\floatstyle{ruled}
\newfloat{listing}{tb}{lst}{}
\floatname{listing}{Listing}
\usepackage{cite}
\usepackage{mathrsfs}
\usepackage{amsfonts}
\usepackage{amsmath}
\usepackage{booktabs}
\usepackage{tabularx}
\newcommand{\tabincell}[2]{\begin{tabular}{@{}#1@{}}#2\end{tabular}}
\usepackage{multirow}
\usepackage{subfigure}
\usepackage{xcolor}

\usepackage{cite}
\usepackage{mathrsfs}
\usepackage{amsfonts}
\usepackage{amsmath}
\usepackage{booktabs}
\usepackage{tabularx}
\usepackage{multirow}
\usepackage{subfigure}
\usepackage{xcolor}
\usepackage{algorithm}
\usepackage{url}
\usepackage{algorithmic}
\usepackage{marvosym}

\input{math_commands}
\nocopyright 

\title{Learning Real-World Image De-Weathering with Imperfect Supervision}
\author{
    Xiaohui Liu\textsuperscript{\rm 1},
    Zhilu Zhang\textsuperscript{\rm 1},
    Xiaohe Wu\textsuperscript{\rm 1 (\Letter)},
    Chaoyu Feng,
    Xiaotao Wang,
    Lei Lei,\\
    Wangmeng Zuo\textsuperscript{\rm 1}
}
\affiliations{
    \textsuperscript{\rm 1}Harbin Institute of Technology\\
    lxh720199@gmail.com,
    cszlzhang@outlook.com,
    xhwu.cpsl.hit@gmail.com,
    wmzuo@hit.edu.cn
}

\usepackage{bibentry}

\begin{document}

\maketitle

\begin{abstract}
Real-world image de-weathering aims at removing various undesirable weather-related artifacts. Owing to the impossibility of capturing image pairs concurrently, existing real-world de-weathering datasets often exhibit inconsistent illumination, position, and textures between the ground-truth images and the input degraded images, resulting in imperfect supervision. Such non-ideal supervision negatively affects the training process of learning-based de-weathering methods. In this work, we attempt to address the problem with a unified solution for various inconsistencies. Specifically, inspired by information bottleneck theory, we first develop a Consistent Label Constructor (CLC) to generate a pseudo-label as consistent as possible with the input degraded image while removing most weather-related degradation. In particular, multiple adjacent frames of the current input are also fed into CLC to enhance the pseudo-label. Then we combine the original imperfect labels and pseudo-labels to jointly supervise the de-weathering model by the proposed Information Allocation Strategy (IAS). During testing, only the de-weathering model is used for inference. Experiments on two real-world de-weathering datasets show that our method helps existing de-weathering models achieve better performance. Codes are available at \url{https://github.com/1180300419/imperfect-deweathering}.
\end{abstract}

\section{Introduction}

\begin{figure*}[!ht]
    \centering
        \centering
        \includegraphics[width=0.95\linewidth]{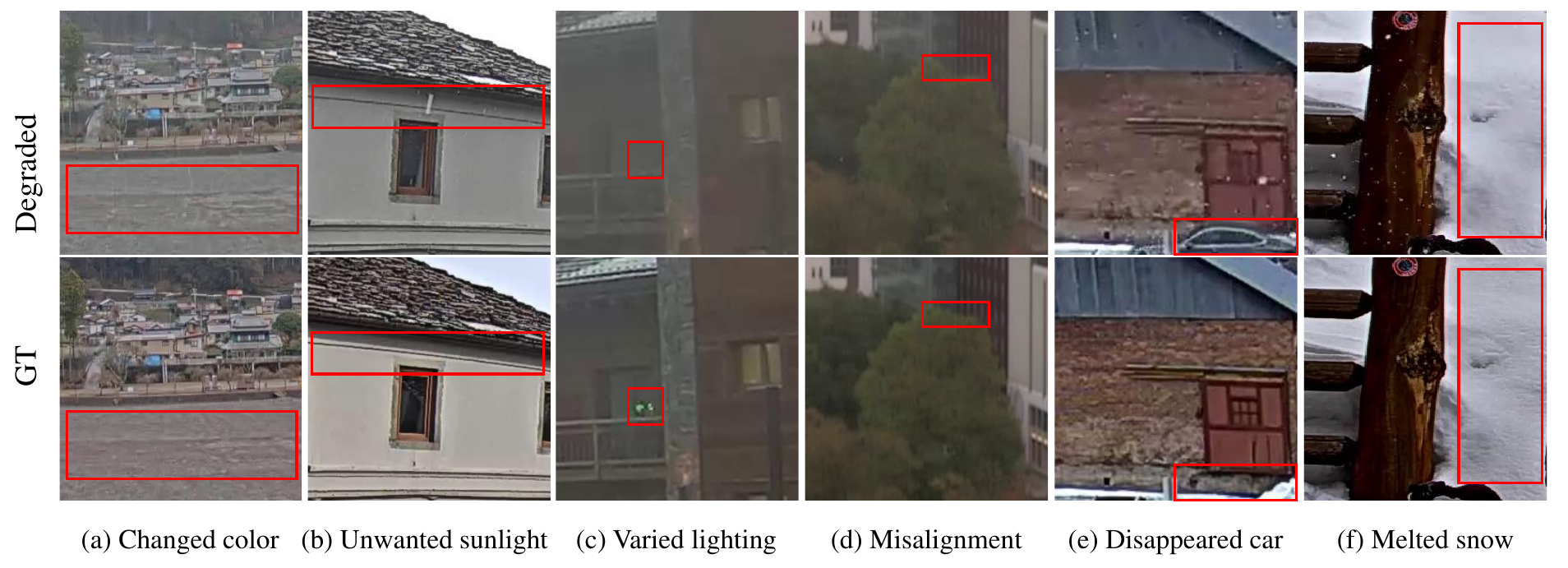}
    \caption{\textbf{Examples of various inconsistent types between the degraded (top) and original GT (bottom) images}. These cases come from two real-world de-weathering datasets (\textit{i.e.}, GT-Rain-Snow and WeatherStream). The pairs in (a)$\sim$(c) show inconsistencies in color and illumination. (d) and (e) show that the pairs are not aligned owing to object movement. (f) shows inconsistent textures due because of snow melting. 
        The inconsistent parts are highlighted in red boxes.}
    \label{fig:dataset_problems}
\end{figure*}

The effectiveness of outdoor computer vision systems \emph{e.g.} monitoring systems and autonomous driving is inevitably impacted by weather conditions.
Image de-weathering, which aims at removing undesirable image degradation caused by adverse weather, \emph{e.g.} rain, fog, and snow, is significant in improving the performance of the subsequent tasks.
Recently, data-driven learning-based de-weathering methods have experienced notable progress.
However, capturing ideal real-world pairs for training neural networks is nearly impossible since it is not feasible to capture a scene simultaneously with and without weather-related artifacts.

To circumvent this problem, some existing de-weathering methods~\cite{valanarasu2022transweather, wang2023smartassign} initially construct paired data by simulating weather degradation and subsequently undergo supervised learning. 
However, synthesized degraded images  have limited realism in modeling complex and variable weather characteristics (\emph{e.g.} raindrop shapes and fog concentration). 
Consequently, models trained using synthetic pairs typically face challenges in effectively generalizing to realistic severe weather scenarios, even leading to artifacts and the loss of original details. 
Other methods~\cite{wu2021contrastive, wei2021deraincyclegan} address the lack of real-world pairs by employing semi-supervised or unsupervised manners. 
While some of these methods~\cite{rai2022fluid, huang2021memory} have demonstrated promising results, challenges remain in separating intricate overlaps of foreground and background information in real-world degraded images~\cite{li2022toward}. 

Recognizing the aforementioned limitations, the recent works~\cite{ba2022not, zhang2023weatherstream} pin their hopes on acquiring real-world image pairs.
They relax the strict requirement for simultaneous acquisition of ideal pairs and instead retrieve degraded and clean images from live streams of landscape scenes on YouTube within the shortest possible time window.
And the conclusion shows that such a manner indeed bridges the domain gap better than synthetic~\cite{hu2019depth, li2019heavy} and semi-real~\cite{wang2019spatial} datasets.
It is worth noting that, despite the adoption of strict filtering conditions during dataset collection to ensure spatial position and illumination consistency between the ground-truth (GT) and degraded images, certain inconsistencies persist, which will have a detrimental impact on the learning process of the de-weathering model.

Actually, the inconsistencies also occur in some other low-level vision  (\emph{e.g.} super-resolution~\cite{zhang2022self,wang2023benchmark}, deblurring~\cite{li2023learning}, learnable ISP~\cite{zhang2021learning}, and etc~\cite{feng2023generating}) tasks.
And their works usually utilize  optical flow estimation and color correction approaches~\cite{he2012guided} for aligning the network output and GT in position and color, respectively.
However, these solutions are less effective when being applied to de-weathering due to the complexity  and diversity of inconsistent types (see Fig.~\ref{fig:dataset_problems}).

In this work, we suggest a novel perspective to alleviate the adverse impact of imperfect supervision in real-world de-weathering datasets, and propose a unified solution for a variety of inconsistencies.
The main motivation is from the information bottleneck (IB) theory~\cite{tishby2015deep}.
Therein, during network training, it has been observed that the mutual information between the learned features and target increases monotonically, while that between the learned features and input increases first and then decreases gradually.
Taking advantage of it, we can develop a \textbf{C}onsistent \textbf{L}abel \textbf{C}onstructor (CLC) for generating a pseudo-label whose non-weather content is as consistent as possible with the input degraded image while degradation are mostly removed.
Considering the characteristics of the dataset itself, multiple adjacent frames of the current input can be also fed into CLC to enhance the ability of the pseudo-label.
Noted that although the pseudo label is better than the original imperfect label in terms of consistency preservation, it may not be as satisfactory as the original label in terms of degradation removal.
Therefore, we combine the two labels and present an \textbf{I}nformation \textbf{A}llocation \textbf{S}trategy (IAS) for supervising the de-weathering model, which draws on the advantages and discards the disadvantages of both.
During inference, only the de-weathering model is used for removing  weather-related degradation.

Experiments are conducted with RainRobust~\cite{ba2022not} and Restormer~\cite{zamir2022restormer} models on GT-Rain-Snow~\cite{ba2022not} and WeatherStream~\cite{zhang2023weatherstream} datasets.
With the proposed method, the quantitative and qualitative results of existing de-weathering models are greatly improved, while not increasing any inference cost.

The main contributions are as follows:
\begin{itemize}
    \item In real-world de-weathering datasets, the content inconsistencies between degraded and labeled images hurt the performance of de-weathering models, and we propose a unified and novel solution for various inconsistencies.
    \item We first feed multiple frames into Consistent Label Constructor (CLC) to generate a pseudo-label consistent with input content, and then combine the pseudo- and original labels for supervising the de-weathering model by the Information Allocation Strategy (IAS).
    \item Experiments demonstrate that our proposed method enhances the performance of existing de-weathering models without incurring additional inference costs.
\end{itemize}

\section{Related Work}

\subsection{De-Weathering Methods}

\noindent\textbf{Deraining.}
Traditional deraining methods~\cite{garg2007vision, chen2013generalized} rely on handcrafted image priors to remove rain streaks. but these priors heavily depend on empirical observations and fail to capture the intrinsic properties of clean images accurately.  In contrast, recent advancements in deep learning-based approaches, utilizing neural convolutional networks (CNNs)~\cite{fu2017removing} or Transformers~\cite{chen2023learning}, have achieved remarkable rain removal performance by learning prior knowledge from paired data. Nonetheless, these methods typically rely on synthetic data and show limited generalization when applied to real-world images. To address this limitation, several studies have explored semi-supervised~\cite{yasarla2021semi} or unsupervised ~\cite{ye2022unsupervised} frameworks for handling real rainy images, given the scarcity of real-world deraining datasets. However, these approaches often face challenges in achieving satisfactory deraining results.

\noindent\textbf{Desnowing.}
Early works usually employ priors derived from rainfall-driven features to model snow particles.
Desnownet~\cite{liu2018desnownet} is the first CNN-based approach introduced for snow removal. 
Recently, DDMSNet~\cite{zhang2021deep} further presents a deep dense multiscale network incorporated semantic and geometric priors to enhance snow removal.
To address the challenges posed by diverse scenes, a multi-scale projection transformer~\cite{10095605} is proposed, which employs a multi-path approach to involve various snow degradation features and utilizes self-attention operation to integrate comprehensive scene context information for clean reconstruction.

\noindent\textbf{Dehazing.}
Previous attempts~\cite{li2019lap} primarily rely on the parameter estimation of the atmospheric scattering model by the handcrafted priors on haze-free images.
With the advent of deep learning techniques, learning-based methods have been developed rapidly by exploring the deployment of CNNs and Transformers effectively~\cite{guo2022image}.
Despite performing excellently well on synthetic data, the above approaches typically suffer from constrained generalization in handling real-world hazy images~\cite{ancuti2021ntire} .
As an alternative, some studies endeavor to employ GANs~\cite{goodfellow2014generative} within the unpaired or unsupervised framework~\cite{chen2022unpaired} for real image dehazing.
However, GANs often generate images with artifacts, which adversely affects the subsequent model training.

\noindent\textbf{Unified De-Weathering.}
To remove various adverse weather conditions via a single set of pre-trained parameters, All-in-One ~\cite{li2020all} deploys multiple encoders associated with different weather types for image restoration. However, it suffers from the model inefficiency problem. To overcome this, TransWeather~\cite{valanarasu2022transweather} presents a transformer-based structure comprising a single encoder-decoder network with learnable weather type queries. In contrast, ~\cite{chen2022learning} draw inspiration from knowledge distillation, and leverage multiple well-trained teacher networks to distill the student model for multiple weather effects. Meanwhile, ~\cite{wang2023smartassign} introduce a multi-task learning way for deraining and desnowing by designing multiple heads. Although the aforementioned works achieve encouraging results in various weather types with a unified framework, they are typically trained on synthetic data, leading to degraded performance in real-world all-weather removal.

\subsection{De-Weathering Datasets}
Single weather removal, such as deraining~\cite{fu2017removing, zhang2018density,  li2019heavy, hu2019depth}, dehazing~\cite{li2018benchmarking, ancuti2019dense, ancuti2021ntire} and desnowing~\cite{liu2018desnownet, chen2020jstasr, chen2021all} has made substantial advancements due to the numerous publicly accessible datasets. 
As for the de-weathering task, due to the absence of dedicated datasets, prevailing methods often explore the integration of existing weather-specific datasets. For example, some works~\cite{valanarasu2022transweather, li2020all} sample images from Snow100K~\cite{liu2018desnownet}, Raindrop~\cite{qian2018attentive} and Outdoor-Rain~\cite{li2019heavy} to facilitate the training of the model.
However, those weather-specific datasets are commonly composed of synthetic paired data given certain degradation.

Recently, SPA~\cite{wang2019spatial} dataset introduces a semi-automatic approach to generate one high-quality clean image from each input sequence of real rain images, and builds the semi-real paired dataset. 
RainRobust~\cite{ba2022not} indeed establishes the real-world paired deraining dataset, \emph{i.e.}, GT-Rain, by enforcing rigorous selection criteria to minimize the environmental variations in time multiplexed pairs.
Unfortunately, it requires significant human and financial resources, which limits the size of the dataset.
To reduce manpower requirements, WeatherStream~\cite{zhang2023weatherstream} then introduce an automatic pipeline capturing real-world weather effects and their clean image pairs, which can be typically regarded as a real-world de-weathering dataset.

\begin{figure}[t]
    \centering
        \centering
        \includegraphics[width=0.98\linewidth]{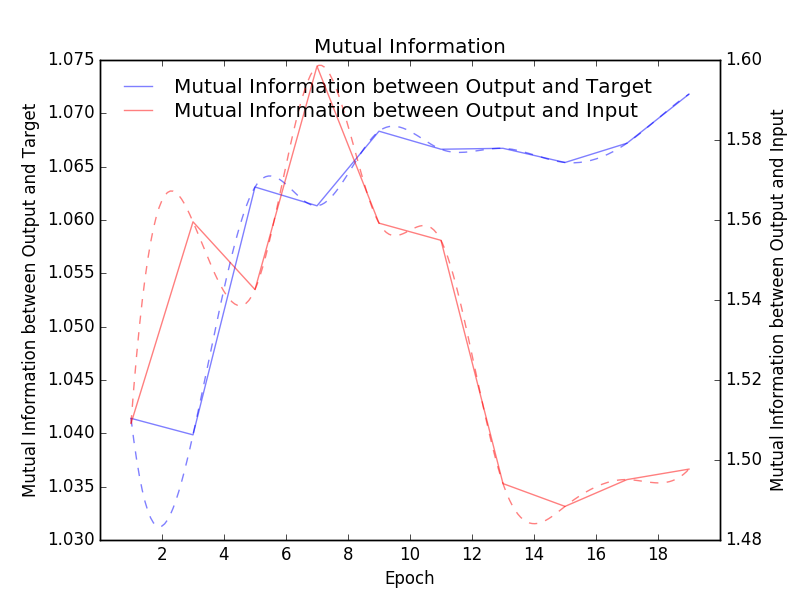}
    \caption{\textbf{Comparison of mutual information in different training periods.} At different training epochs, we test the RainRobust model trained on the GT-Rain-Snow dataset, and measure the mutual information between the output and the target as well as between the output and the input. The model was trained in total $20$ epochs, and we calculate the mutual information every $2$ epochs.}
    \label{fig:mutual_information}
\end{figure}

\section{Proposed Method}
\begin{figure*}[!htbp]
    \centering
        \centering
        \includegraphics[width=0.96\linewidth]{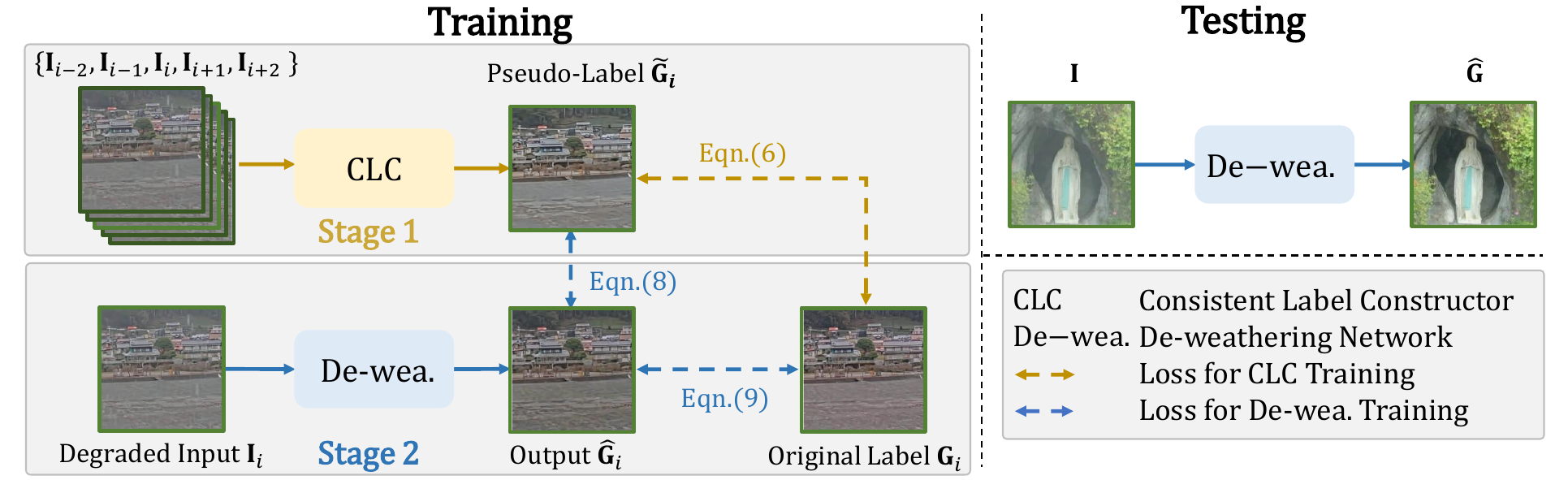}
    \caption{\textbf{Overview of the proposed pipeline}. 
    \textbf{Left: }In training stage $1$, we pre-train a Consistent Label Constructor (CLC), which utilizes multiple degraded frames to produce a pseudo-label $\mathbf{\tilde{G}}_i$ consistent with the input degraded image $\mathbf{I}_i$. 
    In training stage $2$, we leverage both the original label $\mathbf{G}_i$ and pseudo-label $\mathbf{\tilde{G}}_i$  to constrain the output $\mathbf{\hat{G}}_i$ of the de-weathering model.
    \textbf{Right: }During testing, only the de-weathering model is used to remove weather-related degradation.
    Better viewed in color.
    }
  \label{fig:pipeline}
\end{figure*}

\subsection{Motivation and Overall Pipeline}

Denote by $\mathbf{I}$ and $\mathbf{G}$ a degraded image caused by severe weather (\emph{e.g.}, rain, fog, and snow) and the corresponding clean ground-truth (GT) image. 
The learning-based de-weathering model $\mathcal{D}$ predicts output $\mathbf{\hat{G}}$ without weather-related degradation from $\mathbf{I}$, \emph{i.e.},
\begin{equation}
\mathbf{\hat{G}} = \mathcal{D} (\mathbf{I}; \Theta_\mathcal{D}).
\end{equation}
$\Theta_\mathcal{D}$ denotes the model parameters, whose optimization objective can be defined by,
\begin{equation}
    \Theta_\mathcal{D}^\ast = \arg\min_{\Theta_\mathcal{D}} \mathbb{E}_{\mathbf{I},\mathbf{G}} [\mathcal{L_\mathcal{D}}(\mathbf{\hat{G}}, \mathbf{G})],
    \label{eqn:de-wea}
\end{equation}
where $\mathcal{L}_\mathcal{D}$ denotes the loss functions.

Generally, the content of GT image $\mathbf{G}$ should remain the same as that of input  $\mathbf{I}$, except for the weather-related degradation to be removed.
But for real-world de-weathering, $\mathbf{G}$ and $\mathbf{I}$ cannot be captured at the same time, and they have to be collected successively.
During this collection process, changes in weather and scenes are unavoidable and uncontrollable.
As a result, there are inconsistencies in illumination, spatial position, and texture between $\mathbf{G}$ and $\mathbf{I}$.
Although recent works~\cite{ba2022not, zhang2023weatherstream} have set strict filtering and selection conditions to alleviate this problem, some subtle inconsistencies still exist, as shown in Fig.~\ref{fig:dataset_problems}.
Intuitively, these irregular and diverse inconsistencies will inevitably affect the optimization of parameters $\Theta_\mathcal{D}$, harming the performance of the de-weathering model.

It is worth mentioning some other real-world low-level vision  (\emph{e.g.} super-resolution~\cite{zhang2022self,wang2023benchmark}, deblurring~\cite{li2023learning}, learnable ISP~\cite{zhang2021learning}, and etc~\cite{feng2023generating}) tasks face this problem as well, and the corresponding solutions vary.
For inconsistent illumination, guided filter~\cite{he2012guided} is adopted to correct output color to target one, and then calculate the loss value~\cite{wei2020learning}.
For spatial misalignment, optical flow estimation~\cite{sun2018pwc} is widely used for aligning pairs, and some works propose more robust loss functions, \emph{e.g.}, CoBi~\cite{kim2013cobi} and misalignment-tolerate $\ell_1$~\cite{xia2023image} loss.
Regrettably, these single and coarse-grained inconsistency mitigation methods are less effective for real-world de-weathering pairs whose inconsistencies are diverse and fine-grained, as evidenced by the experimental results presented in Sec.\ref{sec:ablation_other}.

In this work, we aim to explore a unified approach for alleviating the adverse impact caused by a variety of inconsistencies in real-world de-weathering datasets.
Inspired by information bottleneck (IB) theory~\cite{tishby2015deep}, we propose a novel perspective for addressing the problem.
Specifically, the IB principle has observed that the mutual information between the network features and target goes up monotonically, while that between the network features and input goes up first and then goes down during the training, as shown in Fig.~\ref{fig:mutual_information}.
In other words, for the target information, the network continuously learns. 
For the input information, the network absorbs first and then eliminates the part that is irrelevant to the target.
Thus, it gives us an opportunity to construct a pseudo-label as consistent as possible with the input. 
Simultaneously, for de-weathering with non-ideal supervision, it may be more difficult to fit weather-irrelevant irregular disturbance from the target than to remove weather-related degradation.
So the pseudo-label is promising to remove most of the degradation while not fitting the imperfect parts of the original GT.

Furthermore, we develop a Consistent Label Constructor (CLC) for generating pseudo-labels.
Multiple neighboring frames of the current degraded image are also fed into CLC to enhance the consistency between the pseudo-label and the degraded image.
The details of CLC are described in Sec.~\ref{sec:CLC}.
Additionally, noted that it is not feasible when only taking the pseudo-label as supervision for the de-weathering model, as the pseudo-label has to make a trade-off between consistency preservation and degradation removal.
Instead, we combine the pseudo-label and original imperfect labels to re-train a de-weathering model by the proposed Information Allocation Strategy (IAS), which is detailed in Sec.~\ref{sec:IAS}.
Please see Fig.~\ref{fig:pipeline} for the overview of the proposed pipeline.

\subsection{Consistent Label Constructor}\label{sec:CLC}

The CLC model $\mathcal{C}$ learns pseudo-label $\mathbf{\tilde{G}}_i$ from input $i$-th degraded image $\mathbf{I}_i$, \emph{i.e.},
\begin{equation}
\mathbf{\tilde{G}}_i = \mathcal{C} (\mathbf{I}_i; \Theta_\mathcal{C}).
\label{eqn:clc_1}
\end{equation}
The parameters $\Theta_\mathcal{C}$ can be optimized by the constraints between the original label $\mathbf{G}_i$ and $\mathbf{\tilde{G}}_i$, which is written as,
\begin{equation}
    \Theta_\mathcal{C}^\ast = \arg\min_{\Theta_\mathcal{C}} \mathbb{E}_{\mathbf{I}_i,\mathbf{G}_i} [\mathcal{L_\mathcal{C}}(\mathbf{\tilde{G}}_i, \mathbf{G}_i)],
    \label{eqn:clc_p}
\end{equation}
where $\mathcal{L}_\mathcal{C}$ denotes the loss functions of CLC model.

Moreover, given that real-world de-weathering datasets are usually collected from live streams, multiple adjacent frames of degraded images can be easily obtained.
Thus, we can naturally feed the neighboring $2n$ frames into the CLC model.
Eqn.~(\ref{eqn:clc_1}) can be modified as,
\begin{equation}
    \mathbf{\tilde{G}}_i=\mathcal{C}(\mathbf{I}_{i-n},
    ...,
    \mathbf{I}_{i-1}, \mathbf{I}_i,  \mathbf{I}_{i+1},
    ...,
    \mathbf{I}_{i+n}; \Theta_\mathcal{C}).
    \label{eqn:clc}
\end{equation}
On the one hand, it does not hinder CLC to produce a pseudo-label consistent with $\mathbf{I}_i$, as the weather-independent content is also consistent between frames captured in a short amount of time.
On the other hand, sometimes complementary multi-frames help CLC to generate a clearer pseudo-label, as the rain streaks and snowflakes cannot always be fixed at the same position in the frame.

For the loss function in Eqn.~(\ref{eqn:clc_p}), we adopt $\ell_1$ as well as structural similarity (SSIM)~\cite{wang2003multiscale} distance between the output $\mathbf{\tilde{G}}_i$ and the original label $\mathbf{G}_i$, which can be written as,
\begin{equation}
    \mathcal{L_\mathcal{C}}(\mathbf{\tilde{G}}_i, \mathbf{G}_i) = ||\mathbf{\tilde{G}}_i - \mathbf{G}_i||_{1} + (1 - \mathcal{L}_{SSIM}(\mathbf{\tilde{G}}_i, \mathbf{G}_i)),
\end{equation}
where  $\mathcal{L}_{SSIM}$ denotes the multi-scale  SSIM function.

The network structure of CLC depends on that of the de-weathering model used in our experiments.
For example, when the de-weathering adopts Restormer~\cite{zhang2023weatherstream},  we simply modified its single encoder to multiple ones as the CLC model.
The details about the CLC architecture can be seen in the suppl.
In addition, we set $n$ to $2$ for experiments.

\subsection{Information Allocation Strategy}\label{sec:IAS}

\begin{table*}[!t]
  \caption{\textbf{Quantitative improvements of de-weathering models when applying the proposed method.} Experiments are conducted with Restormer and RainRobust networks on GT-Rain-Snow and WeatherStream datasets, respectively.}
  \label{tab:result}
  \centering\noindent
  \centering
  \begin{center}
    \begin{tabularx}{0.98\textwidth}{p{0.06\textwidth} p{0.24\textwidth}XXXXXXXX}
        \toprule
        \multicolumn{1}{c}{\multirow{2}{*}{Dataset}}
        & \multicolumn{1}{l}{\multirow{2}{*}{Mothod}} 
        & \multicolumn{2}{c}{Rain}
        & \multicolumn{2}{c}{Fog}
        & \multicolumn{2}{c}{Snow}
        & \multicolumn{2}{c}{Overall}\\
        \cmidrule(l){3-4}
        \cmidrule(l){5-6}
        \cmidrule(l){7-8}
        \cmidrule(l){9-10}
        {} & {} & {PSNR$\uparrow$} & {SSIM$\uparrow$} & {PSNR$\uparrow$} & {SSIM$\uparrow$} & {PSNR$\uparrow$} & {SSIM$\uparrow$} & {PSNR$\uparrow$} & {SSIM$\uparrow$}\\
        \midrule
        \multicolumn{1}{c}{\multirow{4}{*}{\tabincell{c}{GT-Rain-Snow\\~\cite{ba2022not}\\~\cite{zhang2023weatherstream}}}}
        & 
        Restormer~\cite{zamir2022restormer} & 22.63 & 0.7940 & 20.14 & 0.7700 & 21.62 & 0.8123 & 21.51 & 0.7913\\
        &
        \textbf{Ours-Restormer} & \textbf{23.10} & \textbf{0.7971} & \textbf{21.46} & \textbf{0.7860} & \textbf{21.78} & \textbf{0.8178} & \textbf{22.17} & \textbf{0.7995}\\
        \cmidrule(l){2-10}
        & 
        RainRobust~\cite{ba2022not} & 22.83 & 0.7887 & 20.95 & 0.7691 & 22.17 & 0.8058 & 22.01 & 0.7871 \\
        & 
        \textbf{Ours-RainRobust} & \textbf{23.24} & \textbf{0.7980} & \textbf{21.52} & \textbf{0.7860} & \textbf{22.31} & \textbf{0.8107} & \textbf{22.40} & \textbf{0.7977}\\
        \midrule
        \multicolumn{1}{c}{\multirow{4}{*}{\tabincell{c}{WeatherStream\\~\cite{zhang2023weatherstream}}}}
         & 
         Restormer~\cite{zamir2022restormer} & 23.50 & 0.7990 & 22.89 & 0.8034 & 22.29 & 0.8175 & 22.94 & 0.8059\\
        &
        \textbf{Ours-Restormer} & \textbf{23.64} & \textbf{0.8055} & \textbf{23.01} & \textbf{0.8062} & \textbf{22.66} & \textbf{0.8209} & \textbf{23.15} & \textbf{0.8102}\\
        \cmidrule(l){2-10}
        &
        RainRobust~\cite{ba2022not} & 23.46 & 0.7957 & 22.61 & 0.7962 & 22.11 & 0.8143 & 22.78 & 0.8013\\
        & 
        \textbf{Ours-RainRobust} & \textbf{23.68} & \textbf{0.7987} & \textbf{22.71} & \textbf{0.7971} & \textbf{22.54} & \textbf{0.8163} & \textbf{23.02} & \textbf{0.8033}\\
        \bottomrule
    \end{tabularx}
    \end{center}
  \label{result}
\end{table*}

Although the pseudo-label  $\mathbf{\tilde{G}}_i$ is more consistent with the input $\mathbf{I}_i$ than the original label $\mathbf{G}_i$, it may perform worse on degradation removal.
Thus, It is not appropriate to treat $\mathbf{\tilde{G}}_i$  as the only supervision of the de-weathering model.
Naturally, we can also include $\mathbf{G}_i$ as additional guidance, and combine the advantages of $\mathbf{\tilde{G}}_i$ and $\mathbf{G}_i$ for supervising the de-weathering model.
Thus, the loss function in Eqn.~(\ref{eqn:de-wea}) can be written as,
\begin{equation}
    \label{eq:loss_las}
    \mathcal{L_\mathcal{D}}(\mathbf{\hat{G}}_i, \mathbf{\tilde{G}}_i, \mathbf{G}_i) \! =  \! \mathcal{L}_{pse}(\mathbf{\hat{G}}_i, \mathbf{\tilde{G}}_i)  \! + \! \lambda_{ori}\mathcal{L}_{ori}(\mathbf{\hat{G}}_i, \mathbf{G}_i),
\end{equation}
where $\mathcal{L}_{pse}$ and $\mathcal{L}_{ori}$ denote the optimization objectives with pseudo-label and the original label as supervision, respectively.
$\lambda_{ori}$ denotes a weight coefficient. 
The proper design of $\mathcal{L}_{pse}$ and $\mathcal{L}_{ori}$ is essential, and we propose the following Information Allocation Strategy (IAS) for this.

First, IAS should take advantage of the fact that the pseudo-label is more consistent with the input.
Thus, for $\mathcal{L}_{pse}$, we directly adopt the image-level constraint between the output $\mathbf{\hat{G}}_i$ and the pseudo label $\mathbf{\tilde{G}}_i$, \emph{i.e.},
\begin{equation}
    \mathcal{L}_{pse}(\mathbf{\hat{G}}_i, \mathbf{\tilde{G}}_i) = ||\mathbf{\hat{G}}_i - \mathbf{\tilde{G}}_i||_{1}.
\end{equation}

Second, IAS should efficiently leverage clean information from the original label $\mathbf{G}_i$ while mitigating the impact of its inconsistencies as much as possible.
At this time, image-level or pixel-level loss terms are not suitable.
Instead, we impose feature-level and distribution-level constraints because of their robustness to spatial misalignment and color variations between image pairs.
On the one hand, we adopt rain-robust loss~\cite{ba2022not}, which is a variant of contrastive loss and is calculated between intermediate features of the de-weathering networks.
On the other hand, we utilize Sliced Wasserstein (SW)~\cite{zhang2022self} loss, which is a probability distribution distance between VGG~\cite{simonyan2014very} features.
In short, the optimization objective with the original label can be written as,
\begin{equation}
    \mathcal{L}_{ori} = \mathcal{L}_{Robust} + \lambda_{SW} \mathcal{L}_{SW},
\end{equation}
where $\mathcal{L}_{Robust}$ and $\mathcal{L}_{SW}$ denote rain-robust and SW loss terms, respectively.
$\lambda_{SW}$ is the weight for SW loss and set to $0.08$. $\lambda_{ori}$ in Eqn.~(\ref{eq:loss_las}) is set to $0.1$.
The details and formulas of $\mathcal{L}_{Robust}$  and $\mathcal{L}_{SW}$ are shown in the suppl.

\section{Experiments}

\begin{figure*}[!ht]
    \centering
        \centering
        \includegraphics[width=0.95\linewidth]{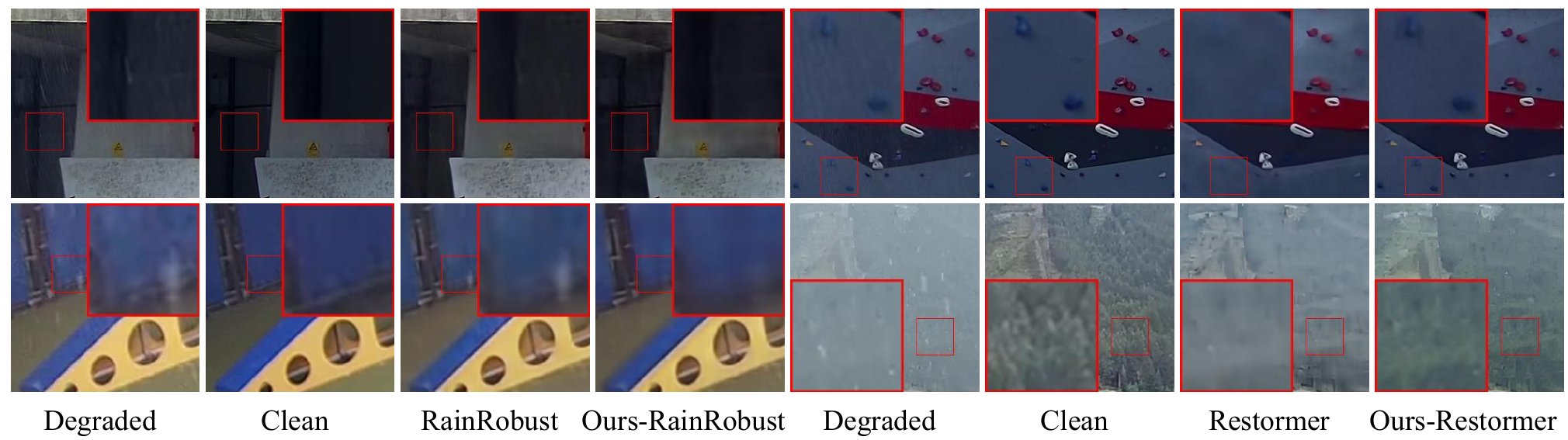}
    \caption{\textbf{Qualitative testing results of the de-weathering models trained with GT-Rain-Snow dataset.}}
  \label{fig:gtrain_results}
\end{figure*}

\begin{figure*}[!ht]
    \centering
        \centering
        \includegraphics[width=0.94\linewidth]{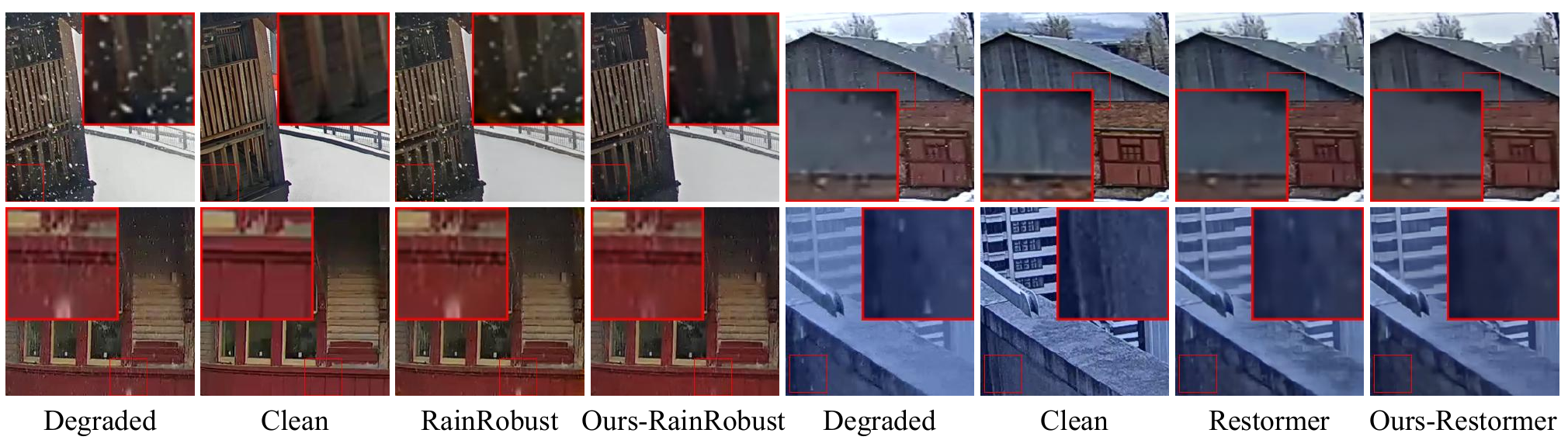}
    \caption{\textbf{Qualitative testing results of the de-weathering models trained with WeatherStream dataset.}}
  \label{fig:weather_results}
\end{figure*}

\begin{figure}[!ht]
    \centering
        \centering
        \includegraphics[width=0.94\linewidth]{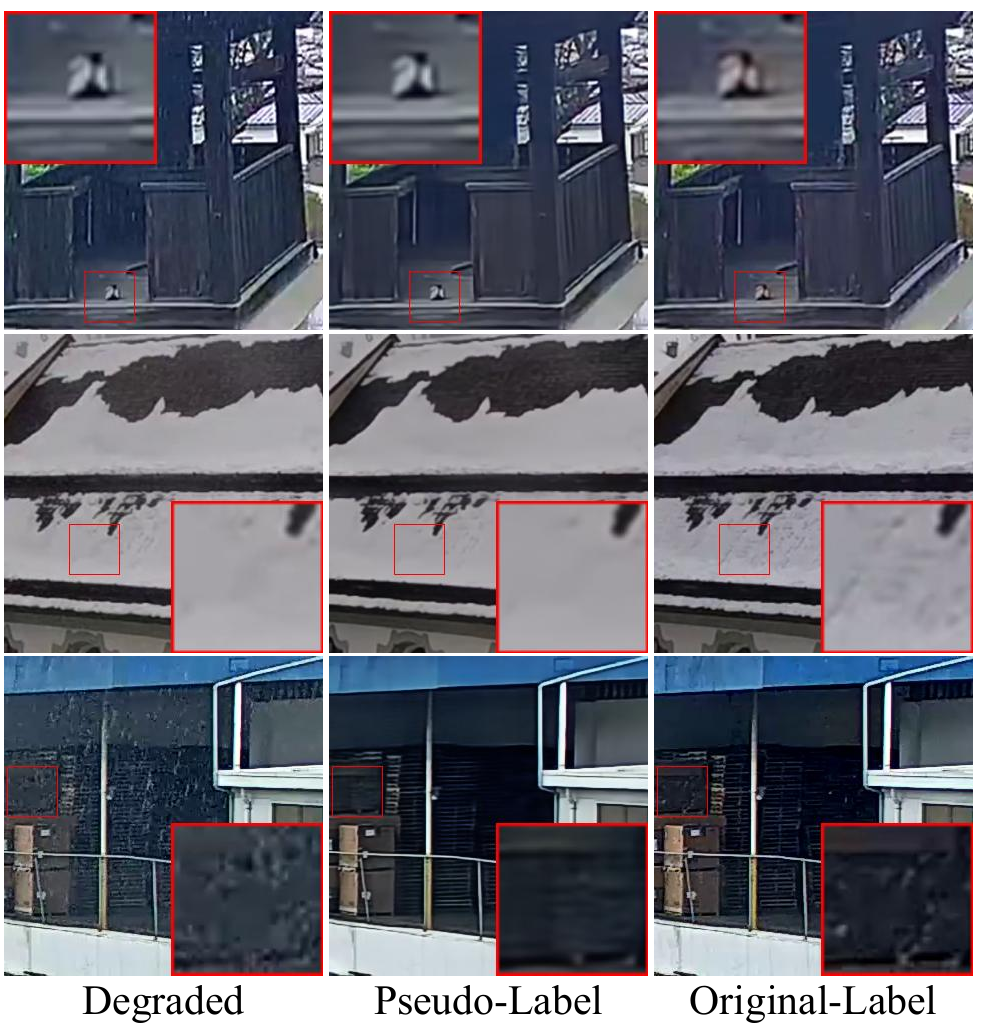}
    \caption{\textbf{Visualizations of pseudo-labels.} }
    \label{fig:pseudo-label}
\end{figure}

\subsection{Datasets}
~\cite{zhang2023weatherstream} offers datasets constructed in two ways. 
One of which, named \textit{GT-Rain-Snow}, follows the protocol of ~\cite{ba2022not} and extends to multiple weather types by manual collection. 
The other \textit{WeatherStream} dataset is obtained by automatic collection.
Specifically, GT-Rain-Snow comprises $129$ scenes and WeatherStream contains a larger set of $424$ scenes. 
Each scene consists of approximately $300$ aligned degraded images and one clean image.
We conduct experiments with the two training datasets, respectively. 
And experiments are all evaluated on the testing dataset proposed by~\cite{zhang2023weatherstream}, which consists of $45$ scenes and covers the sub-tasks of deraining, dehazing and desnowing.

\subsection{Implementation Details}
We integrate our method into the CNN-based RainRobust~\cite{ba2022not} and transformer-based Restormer~\cite{zamir2022restormer} to validate the effectiveness.
During training, we follow~\cite{ba2022not} to employ various augmentation techniques, including random rotation, padding, cropping, Rain-Mix~\cite{guo2021efficientderain} and Snow-Mix~\cite{zhang2023weatherstream}.
For optimizing the models, we adopt the Adam with $\beta_1 = 0.9$ and $\beta_2 = 0.999$.
A warm-up strategy is employed to gradually increase the learning rate from $5\times10^{-5}$ to $2\times10^{-4}$, followed by the cosine annealing strategy to decrease the learning rate from $2\times10^{-4}$ to $10^{-6}$.
Experiments based on RainRobust are conducted on a single GPU. 
The corresponding patch size is $256\times256$, and the batch size is $8$.
%
%
Experiments based on Restormer are conducted on two GPUs.
The corresponding patch and batch sizes are set to $168\times168$ and $6$, respectively.
All experiments are implemented with the PyTorch framework.

\subsection{Comparison with State-of-the-Arts}
\paragraph{Quantitative Comparisons.}
In Table~\ref{tab:result}, we report the PSNR and SSIM of the models trained with GT-Rain-Snow and WeatherStream datasets, respectively.
It can be seen that our method consistently improves the performance of baseline de-weathering models regardless of whether a CNN or Transformer backbone, thereby demonstrating its satisfactory capability.
In particular, when training models with GT-Rain-Snow dataset, our transformer-based Restormer obtains $0.66$dB PSNR improvements while CNN-based RainRobust achieves $0.39$dB PSNR gains.
Although the inconsistencies in the WeatherStream are significantly reduced in comparison with GT-Rain-Snow.
When training models with WeatherStream, our Restormer and RainRobust still obtain $0.21$dB and $0.24$dB PSNR gains, respectively.
Moreover, our method universally boosts the capability of de-weathering across different weather conditions.

\paragraph{Qualitative Comparisons.}
Fig.~\ref{fig:gtrain_results} and Fig.~\ref{fig:weather_results} show visual comparisons on various scenarios.
Our method brings visible improvements with fewer degradation and more consistent color.
For instance, in the bottom-right example of Fig.~\ref{fig:gtrain_results}, our method promisingly recovers the color of heavily foggy images, while Restormer only removes some streaks.
In the top-left example of Fig.~\ref{fig:weather_results}, our de-weathered image contains fewer snow streaks than Rain-Robust.

\subsection{Visualization of Pseudo-Labels}

As mentioned above, our pseudo-labels can  have greater consistency with the degraded images compared to the original label, encompassing color, position, and other aspects. 
Here we give some visual examples in Fig.~\ref{fig:pseudo-label}.
Therein, the first two rows showcase the superior color and texture consistency of pseudo-labels with the degraded images, respectively.
Notably, the utilization of multiple frames in CLC enhances the visual quality in the last row, with pseudo-labels exhibiting more intricate textures, while the original label still retains rain streaks. 

\section{Ablation study}

\subsection{Comparison with Other Inconsistency- Handling Methods} 
\label{sec:ablation_other}

\begin{table}[t]
  \caption{\textbf{Quantitative results when comparing with other inconsistency-handling methods.} The baseline represents the model trained directly with the original label. The results on three sub-tasks are attached in the suppl.}
  \label{tab:ablation-methods}
  \centering
  \begin{tabularx}{0.45\textwidth}{>{\centering\arraybackslash}p{0.25\textwidth}XX}
    \toprule
    \multicolumn{1}{l}{Methods} 
    & {PSNR$\uparrow$} & {SSIM$\uparrow$}\\
     \midrule
    \multicolumn{1}{l}{Baseline} & 22.01 & 0.7871 \\
    \midrule
    \multicolumn{1}{l}{Guided Filter} & 22.14 & 0.7959\\ 
    \multicolumn{1}{l}{Optical Flow} & 22.15 & 0.7932\\
    \multicolumn{1}{l}{Misalign-Tolerate $\ell_1$} & 22.18 & 0.7951\\
    \midrule
    \multicolumn{1}{l}{Guided Filter + Optical Flow} & 22.21 & 0.7935\\
    \multicolumn{1}{l}{Guided Filter + Misalign-Tolerate $\ell_1$} & 22.24 & 0.7976\\
    \midrule
    \multicolumn{1}{l}{Ours} & \textbf{22.40} & \textbf{0.7977}\\
    \bottomrule
  \end{tabularx}
\end{table}

Here we compare our method with alternative ones from other works that can also alleviate the effect of inconsistencies between input and GT images.
We select two types of methods that are robust to color-inconsistent supervision and misaligned supervision, respectively. 
Among them, guided filter~\cite{he2012guided} is adopted for aligning the color between the output and GT~\cite{wei2020learning,wang2023benchmark}.
The optical flow~\cite{sun2018pwc} is used in  some works~\cite{zhang2021learning,li2023learning} for aligning the content spatially.
And the misalignment-tolerate $\ell_1$~\cite{xia2023image} loss is also present to tackle the misalignment.

During experiments, we train the de-weathering model with the single or combined methods by following their ways.
The results are shown in Table~\ref{tab:ablation-methods}.
%
From the table, although these methods can also improve performance compared to the baseline, our method achieves better performance than them.
Furthermore, even combining two complementary techniques (\emph{e.g.}, Guided Filter + Optical Flow), our method still performs better than them.

\subsection{Effect of Number of Frames in CLC}
\begin{table}[t]
    \caption{\textbf{Effect of feeding the different number of frames into the CLC model.} The results of the CLC and corresponding de-weathering models are both reported.}
    \label{tab:ablation-frame}
    \centering
    \begin{tabularx}{0.45\textwidth}{>{\centering\arraybackslash}p{0.08\textwidth}XXXX}
        \toprule
        \multicolumn{1}{c}{\multirow{2}{*}{Frame Number}} 
        & \multicolumn{2}{c}{CLC}
        & \multicolumn{2}{c}{De-Weathering}\\
        \cmidrule(l){2-3}
        \cmidrule(l){4-5}
        {} & {PSNR$\uparrow$} & {SSIM$\uparrow$}
        & {PSNR$\uparrow$} & {SSIM$\uparrow$}\\
        \midrule
        1 & 22.01 & 0.7871 & 22.19 & 0.7971\\
        3 & 22.84 & 0.8199 & 22.34 & 0.7972\\
        5 & 23.30 & 0.8282 & \textbf{22.40} & \textbf{0.7977}\\
        7 & 23.20 & 0.8277 & 22.39 & 0.7976\\ 
        \bottomrule
    \end{tabularx}
\end{table}
\begin{table}[t!]
    \centering
    \caption{\textbf{Quantitative results of  de-weathering model trained with different supervisions.}}
    \centering
    \begin{tabular}{cccc}
        \toprule
        \multicolumn{1}{c}{Pseudo-Label} & \multicolumn{1}{c}{Original Label} & {PSNR$\uparrow$} & {SSIM$\uparrow$} \\
        \midrule
        \multicolumn{1}{c}{$\times$} & \multicolumn{1}{c}{\checkmark} & 22.01 & 0.7871 \\
        \checkmark & $\times$& 22.18 & 0.7903 \\
        \checkmark & \checkmark & \textbf{22.40} & \textbf{0.7977} \\
        \bottomrule
    \end{tabular}
    \centering
    \label{tab:loss_component}
\end{table}

Here we investigate the effect of feeding different number of frames into CLC.
Table~\ref{tab:ablation-frame} evaluates the performance of both CLC and its enhanced de-weathering model.
It can be observed that their performance generally increases as the number of frames increases.
When the number of frames is larger than 5, the performance tends to be saturated.
See suppl. for visual examples.
%

\subsection{Effect of Supervision Setting}
Here we conduct an ablation experiment to investigate the impact of different supervisions when training the de-weathering model.
As illustrated in Table~\ref{tab:loss_component}, benefiting from CLC's meticulous design, taking pseudo-labels for supervising the de-weathering model is more effective than taking the original labels for that.
Nevertheless, combining the two labels can achieve better performance gain.
This result also indicates the complementary nature of the two supervisions, and shows the effectiveness of our proposed IAS.

\section{Conclusion}
In this work, we propose a unified method to mitigate adverse effects of inconsistencies in real-world de-weathering pairs. Firstly, we develop a CLC that generates a pseudo-label consistent with the input degraded image while eliminating most of weather-related degradation. To further enhance the pseudo-label, we incorporate multiple adjacent frames of the input into the CLC. Then we propose an IAS to combine the original imperfect labels and pseudo-labels, thereby providing comprehensive and complementary supervision for the de-weathering networks. Experimental results on two real-world de-weathering datasets demonstrate that our method significantly improves the performance of existing de-weathering models.

\section*{Acknowledgements}
This work was supported in part by the National Natural Science Foundation of China (NSFC) under Grant No. U19A2073 and Heilongjiang Science and Technology Project under Grant 2022ZX01A21.

\appendix

\renewcommand{\thesection}{\Alph{section}}
\renewcommand{\thetable}{\Alph{table}}
\renewcommand{\thefigure}{\Alph{figure}}
\renewcommand{\thealgorithm}{\Alph{algorithm}}

\setcounter{section}{0}
\setcounter{table}{0}
\setcounter{figure}{0}
\setcounter{algorithm}{0}

\twocolumn[
\begin{@twocolumnfalse}
	\section*{\centering{\LARGE Supplementary Material\\[30pt]}}
\end{@twocolumnfalse}
]

\section{Content}
The content of this supplementary material involves:
\begin{itemize}
    \item Details of Sliced Wasserstein (SW) loss in Sec.~\ref{sec:sw_loss}.
    \item Details of rain-robust loss in Sec.~\ref{sec:rain_robust_loss}.
    \item Architecture of Consistent Label Constructor (CLC) in Sec.~\ref{sec:clc_suppl}.
    \item Detailed results of the ablation study in Sec.~\ref{sec:detailed_table}.
    \item More visual results of pseudo-labels in Sec.~\ref{sec:pseudo-label}
    %
    %
    \item More visual comparisons of de-weathering in Sec.~\ref{sec:visual_results}
    \item Out-of-distribution (OOD) results in Sec.~\ref{sec:ood_results}
    
\end{itemize}

\section{Sliced Wasserstein Loss}\label{sec:sw_loss}

As shown in Alg.~\ref{alg:sw_loss}, to calculate the Sliced Wasserstein (SW) loss between the output image $\mathbf{\hat{G}}$ and the target image $\mathbf{G}$, we first get the $2$-dimensional features from VGG19~\cite{simonyan2014very}, following by reshaping them into $1$-dimensional through random liner projection.
Then, we calculate the Wasserstein distance between the output and the target $1$-dimensional probability distributions, which is defined as the element-wise $\ell_1$ distance over sorted $1$-dimensional distributions.

\section{Rain-Robust Loss}\label{sec:rain_robust_loss}

Rain-Robust loss is proposed by~\cite{ba2022not}, which is inspired by the advance in contrastive learning~\cite{chen2020simple}. 
The details of Rain-Robust loss are shown in Alg.~\ref{alg:rain_robust_loss}.
For each image in $\{(\mathbf{I}_i, \mathbf{G}_i)\}_{i=1}^N$, we first utilize a feature extractor to extract features.
After that, we follow InfoNCE~\cite{oord2018representation} criterion to measure the Rain-Robust loss for every positive pair.
Finally, we return the mean value of the Rain-Robust loss for $N$ image pairs.
In the training process of the de-weathering network, we use $N$ pairs of degraded-clean images from the same batch as the input for the Rain-Robust loss, and select the encoder $\mathcal{D_E}(\cdot, \Theta_\mathcal{E})$ of de-weathering network $\mathcal{D} (\cdot, \Theta_\mathcal{D})$ as the feature extractor. 
The temperature coefficient $\tau$ is set to $0.25$.

\section{Consistent Label Constructor}\label{sec:clc_suppl}
\begin{figure}[!ht]
    \centering
        \centering
        \includegraphics[width=\linewidth]{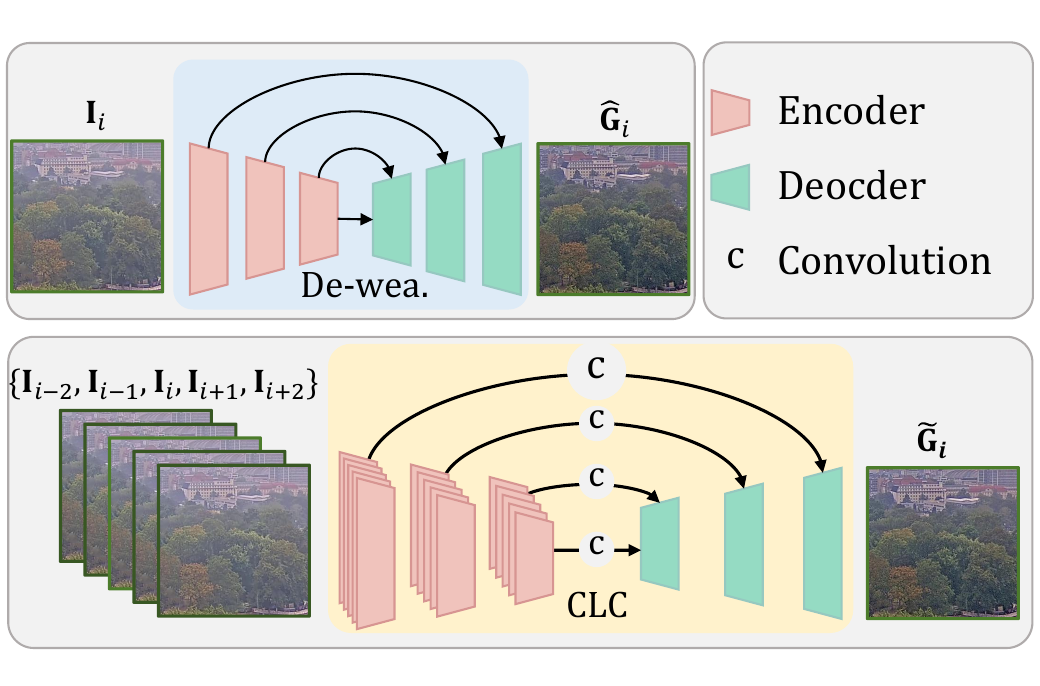}
        \vspace{-6mm}
    \caption{\textbf{The overall architecture of CLC model and 
 the corresponding de-weathering model.}}
  \label{fig:clc}
  \vspace{-2mm}
\end{figure}

Our CLC model corresponds to the de-weathering model, which adopts  Restormer~\cite{zamir2022restormer} and RainRobust ~\cite{ba2022not} in the experiments, respectively.
Both the architectures can be considered as U-Net~\cite{ronneberger2015u} framework, which consists of an encoder (left path) and a decoder (right path). 
As shown in Fig.~\ref{fig:clc}, we maintain the original decoder unchanged, while expanding the single encoder into multiple ones, ensuring that each input frame corresponds to a separate encoder.
Furthermore, to enable the successful integration of the outputs from multiple encoders into the original decoder, we concatenate the outputs together and utilize simple convolutional layers to adjust their channel dimensions.

\section{Detailed Results of the Ablation Study} \label{sec:detailed_table}

In this section, we present supplementary results of the ablation study.
Table~\ref{tab:ablation-other-methods-supp} and Table~\ref{tab:ablation-frame-supp} represent the detailed results of Table $2$ and Table $3$ and in the main text, respectively.

Moreover, we conduct ablation studies of loss terms, as shown in Table~\ref{tab:ablation-lambda}. It brings $0.17$dB PSNR gain when replacing the original label with pseudo-label as supervision, which verifies the effectiveness of CLC model. Coupling with either $\mathcal{L}_{Robust}$ or $\mathcal{L}_{SW}$, the performance can be further enhanced, showing the effectiveness of IAS.

\begin{figure*}[!ht]
    \centering
        \centering
        \includegraphics[width=\linewidth]{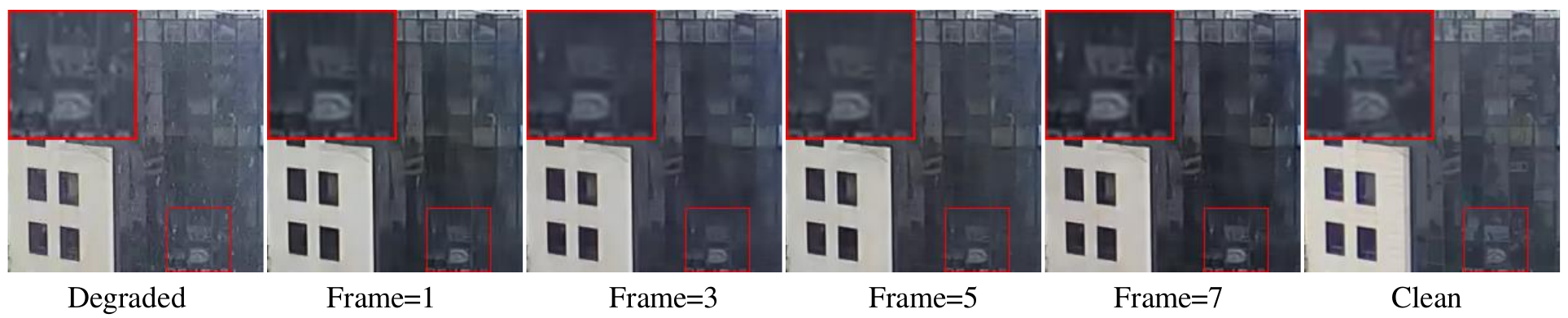}
        \vspace{-6mm}
    \caption{\textbf{Results of the CLC model that takes the different number of frames as input.} }
  \label{fig:teacher_frame_ablation}
  \vspace{-3mm}
\end{figure*}

We also provide visual results of the pseudo-labels that are generated by CLC with different numbers of input frames, as shown in Fig.~\ref{fig:teacher_frame_ablation}. It can be seen that more input frames will improve the visual quality of pseudo-labels, and generate properer texture and color, and fewer artifacts.

\section{Visual Results of Pseudo-Labels} \label{sec:pseudo-label}
\begin{figure*}[!ht]
    \centering
        \centering
    \includegraphics[width=0.8\linewidth]{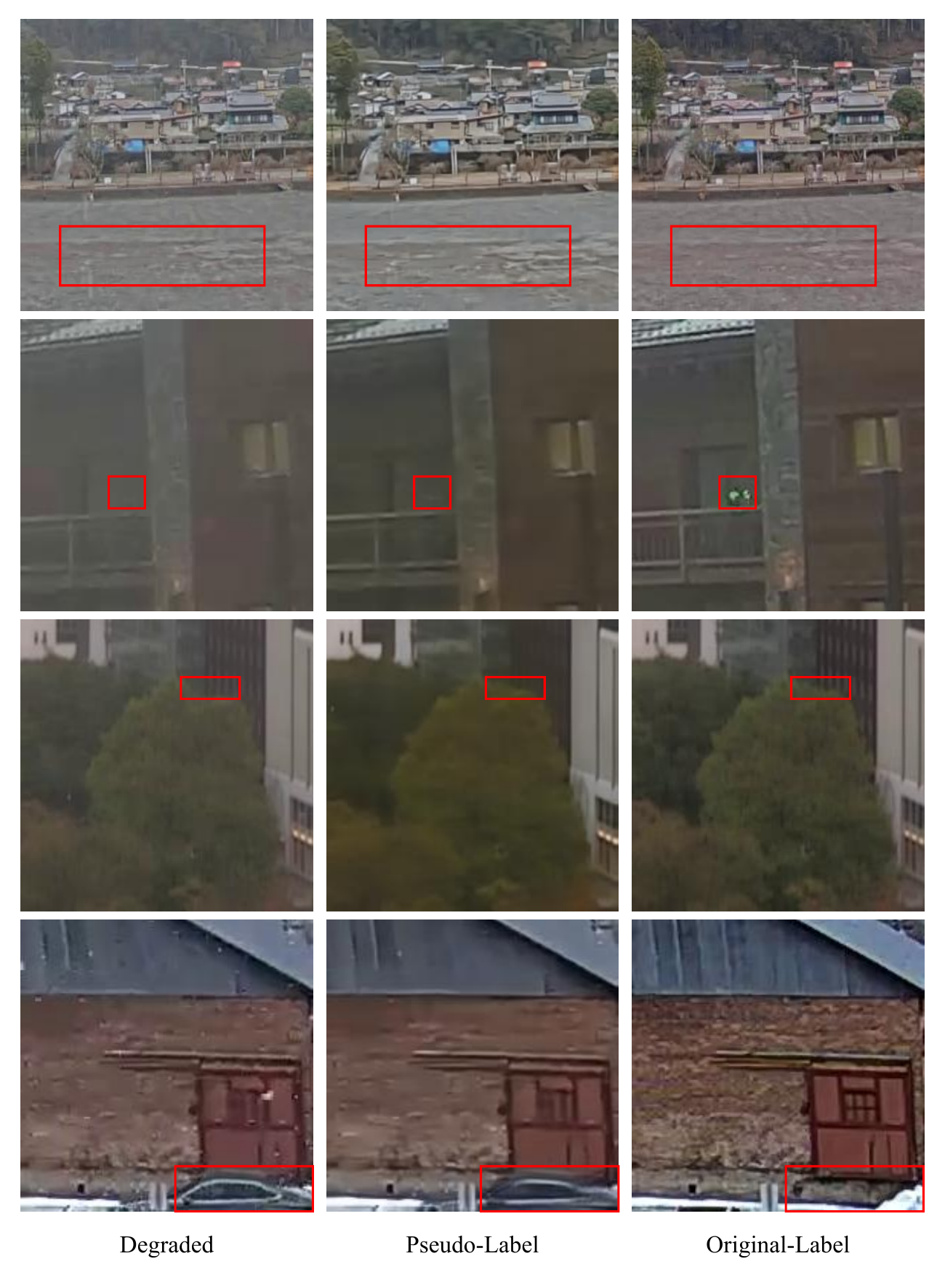}
    \caption{\textbf{Visualizations of pseudo-labels.} Pseudo-labels from top to bottom sequentially show better consistency with the degraded images in color, illumination, spatial position, and moving content.}
  \label{fig:pseudo-labels}
\end{figure*}

In this section, we provide more visual results of pseudo-labels generated by our Consistent Label Constructor (CLC).
As shown in Fig.~\ref{fig:pseudo-labels}, despite the inconsistency between the degraded images and the clean images, our CLC can generate pseudo-labels that exhibit stronger consistency with the degraded images.

\section{Visual Results of De-Weathering} \label{sec:visual_results}
\begin{figure*}[!ht]
    \centering
        \centering
        \includegraphics[width=0.9\linewidth]{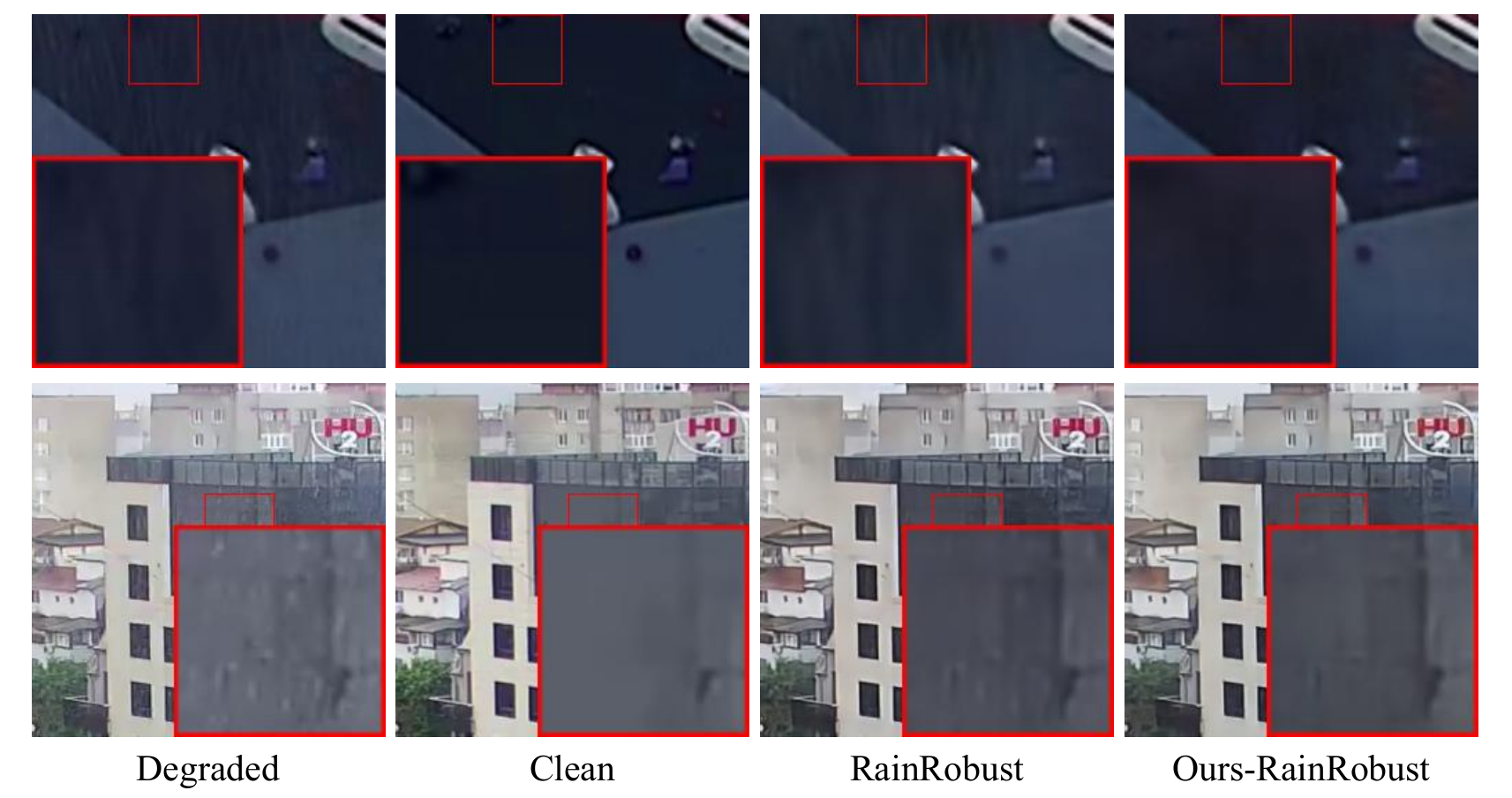}
        \vspace{-2mm}
    \caption{\textbf{Qualitative testing results of the RainRobust model trained with the GT-Rain-Snow dataset.}}
  \label{fig:gtrain_rainrobust}
\end{figure*}

\begin{figure*}[!ht]
    \centering
        \centering
        \includegraphics[width=0.9\linewidth]{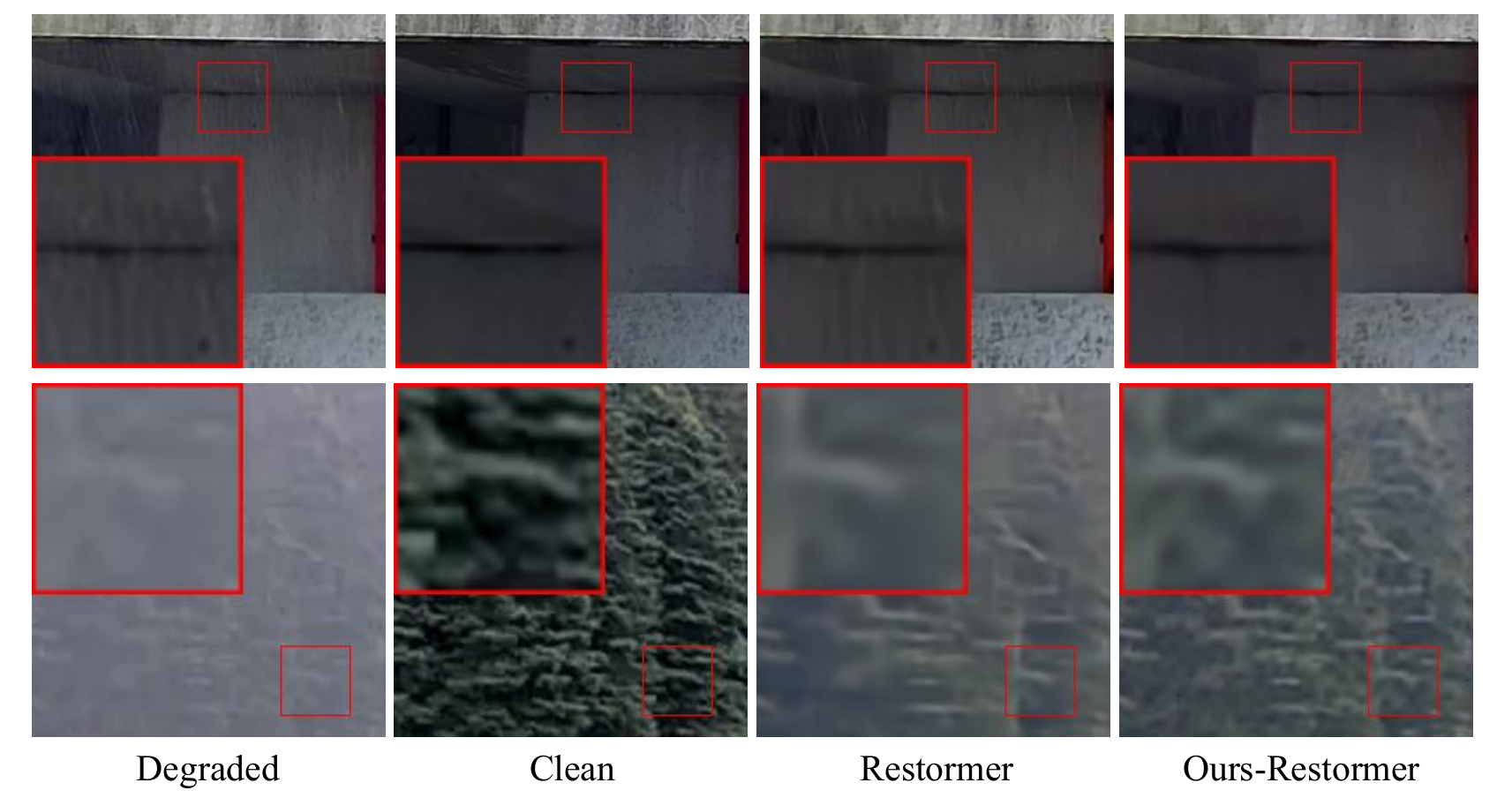}
        \vspace{-2mm}
    \caption{\textbf{Qualitative testing results of the Restormer trained with the GT-Rain-Snow dataset.}}
  \label{fig:gtrain_restormer}
\end{figure*}

\begin{figure*}[!ht]
    \centering
        \centering
        \includegraphics[width=0.9\linewidth]{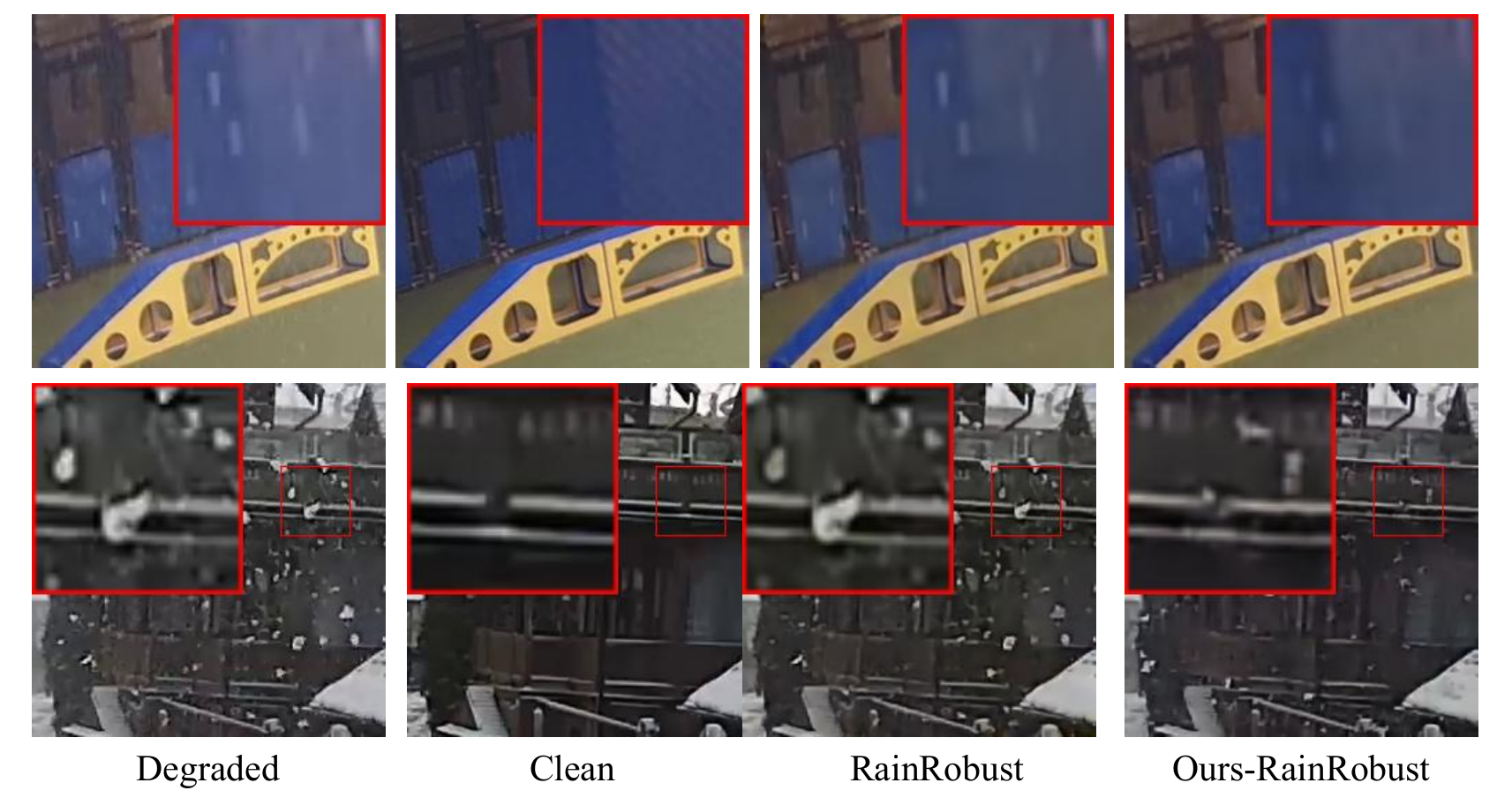}
    \caption{\textbf{Qualitative testing results of the RainRobust model trained with the WeatherStream dataset.}}
  \label{fig:weatherstream_rainrobust}
\end{figure*}

\begin{figure*}[!ht]
    \centering
        \centering
        \includegraphics[width=0.9\linewidth]{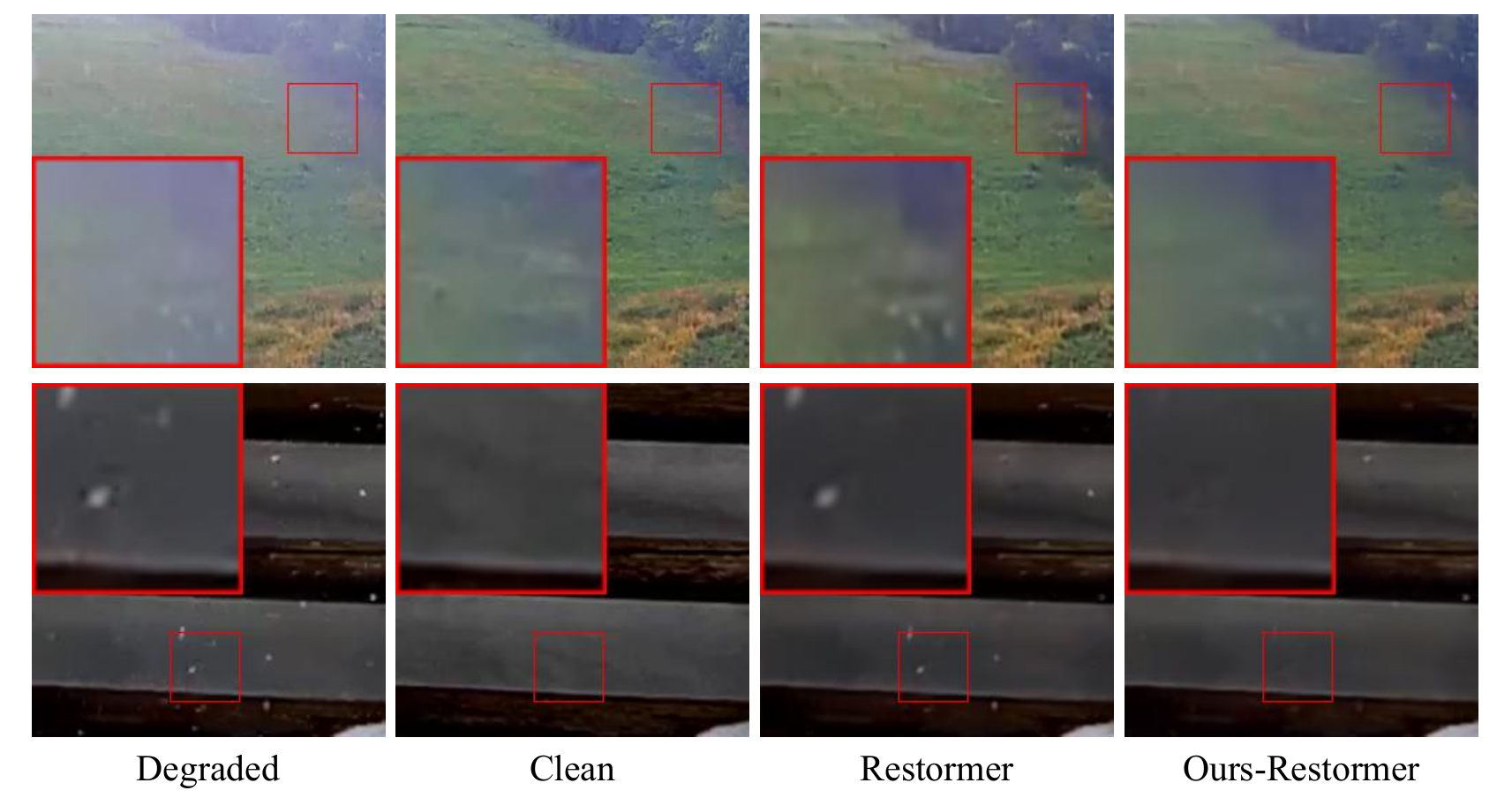}
    \caption{\textbf{Qualitative testing results of the Restormer trained with 
    the WeatherStream dataset.}}
  \label{fig:weatherstream_restormer}
\end{figure*}

In this section, we provide more qualitative comparisons between our proposed method and the baselines.
Fig.~\ref{fig:gtrain_rainrobust} and  Fig.~\ref{fig:gtrain_restormer}  respectively illustrate the results of taking RainRobust and Restormer as de-weathering models when training with the GT-Rain-Snow dataset.
Fig.~\ref{fig:weatherstream_rainrobust} and Fig.~\ref{fig:weatherstream_restormer} respectively illustrate the results of taking RainRobust and Restormer as de-weathering models when training with the WeatherStream dataset. 


\begin{algorithm*}[ht]
\caption{Pseudo code of SW loss}
\begin{algorithmic}[1]
    \REQUIRE $\mathbf{\hat{G}}$: output image, \\
             $\mathbf{G}$: target image, \\
             $\mathbf{M}\in\mathbb{R}^{C' \times C}$: random projection matrix.
    \STATE
        obtain VGG19 features: $\mathbf{U}=\text{VGG}(\mathbf{\hat{G}})$, $\mathbf{V}=\text{VGG}(\mathbf{G})$, $\mathbf{U}$ and $\mathbf{V}$ are both in $\mathbb{R}^{C \times H \times W}$.
    \STATE 
        reshape the feature $\mathbf{U}$ and $\mathbf{V}$ to $\mathbf{U_r}\in\mathbb{R}^{C \times HW}$ and $\mathbf{V_r}\in\mathbb{R}^{C \times HW}$, respectively.
    \STATE
        project the feature $\mathbf{U}$ and $\mathbf{V}$ to $\mathbb{R}^{C' \times HW}$ using $\mathbf{M}$: $\mathbf{U_p}=\mathbf{M}\mathbf{U_f}$, $\mathbf{V_p}=\mathbf{M}\mathbf{V_f}$.
    \STATE
        sort $\mathbf{U_p}$ and $\mathbf{V}_p$: $\mathbf{U_s}=\text{Sort}(\mathbf{U_p}, \text{dim}=1)$, $\mathbf{V_s}=\text{Sort}(\mathbf{V_p}, \text{dim}=1)$.
    \RETURN $\Vert\mathbf{U_s}-\mathbf{V_s}\Vert_1$.
\end{algorithmic}
\label{alg:sw_loss}
\end{algorithm*}

\vspace{1cm}
\begin{algorithm*}[ht]
\caption{Pseudo code of Rain-Robust loss}
\begin{algorithmic}[1]
    \REQUIRE $\{(\mathbf{I}_i, \mathbf{G}_i)\}_{i=1}^N$: $N$ degraded-clean image pairs from different scenes,\\
             $\mathcal{D_E}(\cdot, \Theta_\mathcal{E})$: encoder of the current de-weathering model,\\
             $\tau$: temperature coefficient.
    \FOR {{$i$ from $1$ to $N$}}
        \STATE
            $\mathbf{U}_i=\mathcal{D_E}(\mathbf{I}_i, \Theta_\mathcal{E})$ and $\mathbf{V}_i=\mathcal{D_E}(\mathbf{G}_i, \Theta_\mathcal{E})$,
    \ENDFOR
    \FOR {{$i$ from $1$ to $N$}}
        \STATE
            $\mathcal{L}_{\mathbf{VU}_i} = -\text{log}\frac{\text{exp}(\text{sim}_{\text{cos}}(\mathbf{U}_i, \mathbf{V}_i)/\tau)}{\Sigma_{j=1, j\neq i}^N(\text{exp}(\text{sim}_{\text{cos}}(\mathbf{U}_i, \mathbf{U}_j)/\tau) + \text{exp}(\text{sim}_{\text{cos}}(\mathbf{U}_i, \mathbf{V}_j)/\tau))}$,\\
            $\mathcal{L}_{\mathbf{UV}_i} = -\text{log}\frac{\text{exp}(\text{sim}_{\text{cos}}(\mathbf{V}_i, \mathbf{U}_i)/\tau)}{\Sigma_{j=1, j\neq i}^N(\text{exp}(\text{sim}_{\text{cos}}(\mathbf{V}_i, \mathbf{U}_j)/\tau) + \text{exp}(\text{sim}_{\text{cos}}(\mathbf{V}_i, \mathbf{V}_j)/\tau))}$,
    \ENDFOR
        \RETURN $\frac{\Sigma_{i=1}^N(\mathcal{L}_{\mathbf{VU}_i} + \mathcal{L}_{\mathbf{UV}_i})}{N}$.
\end{algorithmic}
\label{alg:rain_robust_loss}
\end{algorithm*}

\begin{table*}[!ht]
  \vspace{6mm}
  \caption{\textbf{Quantitative results when comparing with other inconsistency-handling methods.} The baseline represents the model trained directly with the original label.}
  \vspace{-1em}
  \label{tab:ablation-other-methods-supp}
  \centering\noindent
  \centering
  \begin{center}
    \begin{tabular}{ccccccccc}
        \toprule
        \multicolumn{1}{c}{\multirow{2}{*}{Methods}} 
        & \multicolumn{2}{c}{Rain}
        & \multicolumn{2}{c}{Fog}
        & \multicolumn{2}{c}{Snow}
        & \multicolumn{2}{c}{Overall}\\
        \cmidrule(l){2-3}
        \cmidrule(l){4-5}
        \cmidrule(l){6-7}
        \cmidrule(l){8-9}
        {} & {PSNR$\uparrow$} & {SSIM$\uparrow$} & {PSNR$\uparrow$} & {SSIM$\uparrow$} & {PSNR$\uparrow$} & {SSIM$\uparrow$} &
        {PSNR$\uparrow$} & {SSIM$\uparrow$}\\
        \midrule
         \multicolumn{1}{l}{Baseline} & 22.83 & 0.7887 & 20.95 & 0.7691 & 22.17 & 0.8058 & 22.01 & 0.7871 \\
        \midrule
        \multicolumn{1}{l}{Guided Filter} & 23.00 & 0.7960 & 20.91 & 0.7791 & 22.43 & 0.8148 & 22.14 & 0.7959\\ 
        \multicolumn{1}{l}{Optical Flow} & 22.81 & 0.7928 & 21.35 & 0.7818 & 22.09 & 0.8065 & 22.15 & 0.7932\\
        \multicolumn{1}{l}{Misalign-Tolerate $\ell_1$} & 22.86 & 0.7919 & 21.26 & 0.7806 & 22.40 & 0.8135 & 22.18 & 0.7951\\
        \midrule
        \multicolumn{1}{l}{Guided Filter + Optical Flow} & 23.06 & 0.7962 & 21.11 & 0.7776 & 22.36 & 0.8080 & 22.21 & 0.7935\\
        \multicolumn{1}{l}{Guided Filter + Misalign-Tolerate $\ell_1$} & 22.89 & 0.7955 & 22.21 & 0.7914 & 21.45 & 0.8072 & 22.24 & 0.7976\\
        \midrule
        \multicolumn{1}{l}{Ours} & 23.24 & 0.7980 & 21.52 & 0.7860 & 22.31 & 0.8107 & 22.40 & 0.7977\\
        \bottomrule
    \end{tabular}
    \end{center}
\end{table*}

\section{Out-of-Distribution (OOD) Results} \label{sec:ood_results}

\begin{figure*}[!h]
    \centering
        \centering
\includegraphics[width=0.9\linewidth]{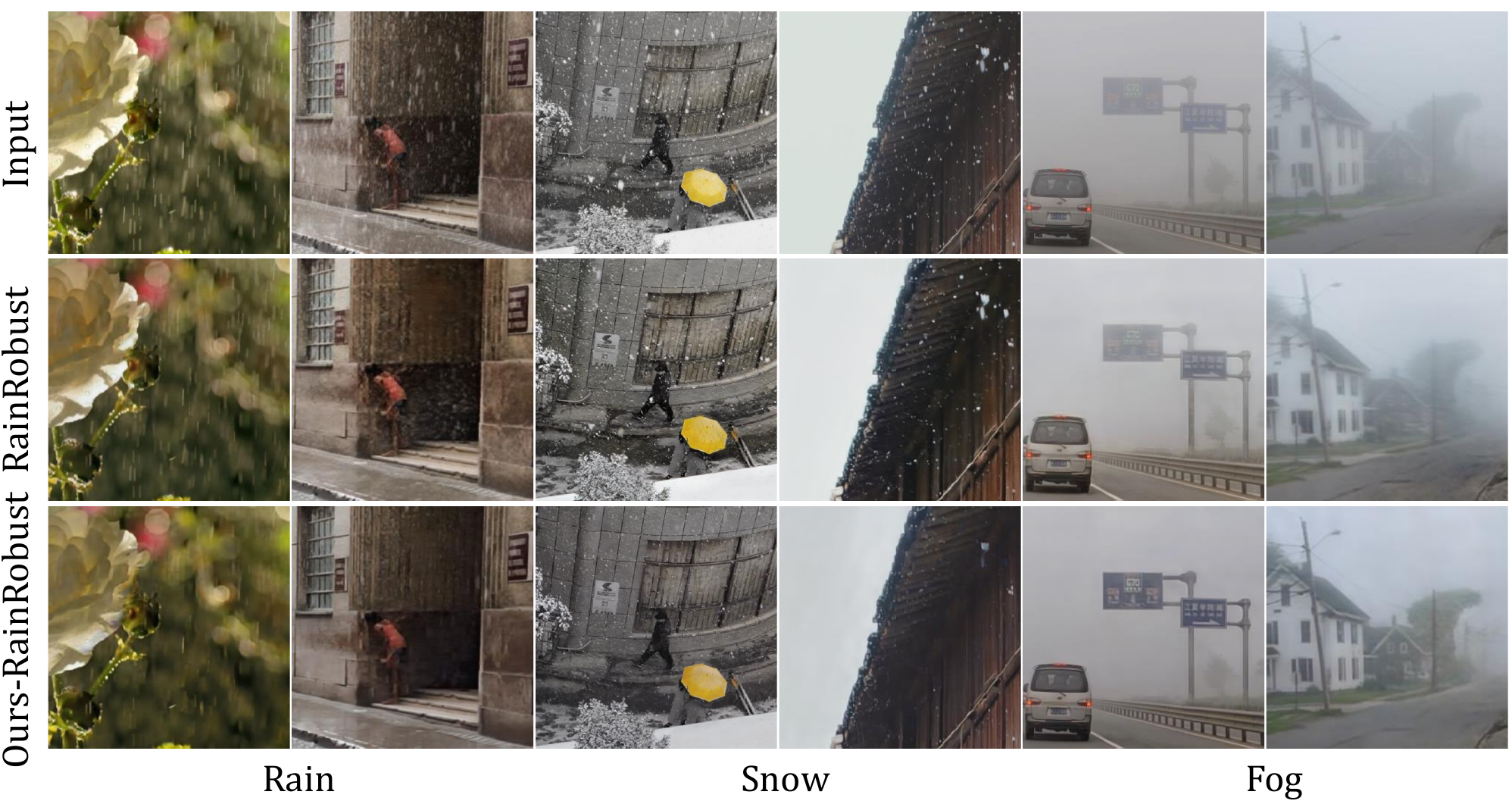}  
    \caption{\textbf{Qualitative comparisons on OOD images.} }
    \label{fig:real_internet}
\end{figure*}

Here we evaluate the method generalizability on three types of severe weather images, including 145 rainy images from the widely-used Internet-Data~\cite{wang2019spatial}, 100 snowy images, and 100 foggy images collected from Google.
Due to the lack of ground-truths, we report two non-reference metrics in Tab.~\ref{tab:ood_result} (\textit{i.e.}, PI~\cite{blau20182018} and CLIP-IQA~\cite{wang2023exploring}). The results on three sub-tasks show that our method still brings promising improvements.
Besides, from Fig.~\ref{fig:real_internet}, our method predicts clearer images and preserves more consistent appearances with the input.

\clearpage

\begin{table*}[!ht]
  \caption{\textbf{Effect of feeding the different number of frames into the CLC model.} The results of the CLC and corresponding de-weathering models on three sub-tasks are both reported.}
  \vspace{-1em}
  \label{tab:ablation-frame-supp}
  \centering\noindent
  \centering
  \begin{center}
    \begin{tabular}{ccccccccccc}
        \toprule
        \multicolumn{1}{c}{\multirow{2}{*}{Frame Number}} 
        & \multicolumn{2}{c}{CLCN}
        & \multicolumn{2}{c}{Rain}
        & \multicolumn{2}{c}{Fog}
        & \multicolumn{2}{c}{Snow}
        & \multicolumn{2}{c}{Overall}\\
        \cmidrule(l){2-3}
        \cmidrule(l){4-5}
        \cmidrule(l){6-7}
        \cmidrule(l){8-9}
        \cmidrule(l){10-11}
        {} & {PSNR$\uparrow$} & {SSIM$\uparrow$} & {PSNR$\uparrow$} & {SSIM$\uparrow$} & {PSNR$\uparrow$} & {SSIM$\uparrow$} & {PSNR$\uparrow$} & {SSIM$\uparrow$} &
        {PSNR$\uparrow$} & {SSIM$\uparrow$}\\
        \midrule
        1 & 22.01 & 0.7871 & 22.73 & 0.7965 & 21.81 & 0.7866 & 21.95 & 0.8095 & 22.19 & 0.7971\\
        3 & 22.83 & 0.8198 & 23.40 & 0.7994 & 21.26 & 0.7852 & 22.19 & 0.8080 & 22.34 & 0.7972\\
        5 & 23.30 & 0.8281 & 23.24 & 0.7980 & 21.52 & 0.7860 & 22.31 & 0.8107 & 22.40 & 0.7977\\
        7 & 23.20 & 0.8276 & 23.25 & 0.7998 & 21.52 & 0.7840 & 22.29 & 0.8101 & 22.39 & 0.7976\\ 
        \bottomrule
    \end{tabular}
    \end{center}
\end{table*}

\begin{table*}[t]
  \caption{\textbf{Quantitative results of  de-weathering model trained with different supervisions.}}
  \vspace{-3mm}
  \label{tab:ablation-lambda}
  \centering\noindent
  \centering
  \begin{center}
    \begin{tabular}{c c c c c c c c c c c c c}
        \toprule
        \multicolumn{2}{c}{\multirow{1}{*}{Original Label}}
        & \multicolumn{1}{c}{\multirow{1}{*}{Pseudo-Label}} 
        & \multicolumn{2}{c}{Rain}
        & \multicolumn{2}{c}{Fog}
        & \multicolumn{2}{c}{Snow}
        & \multicolumn{2}{c}{Overall}\\
        \cmidrule(l){1-2}
        \cmidrule(l){3-3}
        \cmidrule(l){4-5}
        \cmidrule(l){6-7}
        \cmidrule(l){8-9}
        \cmidrule(l){10-11}
        {$\mathcal{L}_{Robust}$} & {$\mathcal{L}_{SW}$} & {$\mathcal{L}_{pse} (\ell_1)$ } & {PSNR$\uparrow$} & {SSIM$\uparrow$} & {PSNR$\uparrow$} & {SSIM$\uparrow$} & {PSNR$\uparrow$} & {SSIM$\uparrow$} & 
        {PSNR$\uparrow$} & {SSIM$\uparrow$}\\
        \midrule
        \checkmark & \checkmark & $\times$ & 22.83 & 0.7887 & 20.95 & 0.7691 & 22.17 & 0.8058 & 22.01 & 0.7871\\
        $\times$ & $\times$ & \checkmark & 22.87 & 0.7883 & 21.93 & 0.7820 & 21.57 & 0.8021 & 22.18 & 0.7903\\
        \midrule
        $\times$ & \checkmark & \checkmark & 23.19 & 0.7977 & 21.34 & 0.7834 & 21.93 & 0.8032 & 22.22 & 0.7946\\
        \checkmark & $\times$ & \checkmark & 23.20 & 0.7996 & 21.31 & 0.7848 & 21.96 & 0.8051 & 22.28 & 0.7963\\
        \midrule
        \checkmark & \checkmark & \checkmark & 23.24 & 0.7980 & 21.52 & 0.7860 & 22.31 & 0.8107 & 22.40 & 0.7977\\
        \bottomrule
    \end{tabular}
    \end{center}
\end{table*}

\clearpage

\begin{table*}[!ht]
  \caption{\textbf{Non-reference metrics tested on OOD images.}}
  \label{tab:ood_result}
  \centering\noindent
  \scalebox{1}
  {
  \begin{tabularx}{0.6\textwidth}{p{0.1\textwidth} p{0.1501\textwidth}XX}
        \toprule
        \multirow{1}{*}{\small Dataset} & \multirow{1}{*}{\small Method} & \multirow{1}{*}{\small PI$\downarrow$} & \multirow{1}{*}{\small  CLIP-IQA$\uparrow$}\\
        \midrule
        \multicolumn{1}{c}{\multirow{4}{*}{\tabincell{c}{GT-Rain-Snow}}}
        & 
        Restormer & 4.714 & 0.3726\\
        &
        \textbf{Ours-Restormer} & $3.905_{\downarrow0.809}$ & $0.3905_{\uparrow0.0179}$\\
        \cmidrule(l){2-4}
        & 
        RainRobust & 3.801 & 0.3380\\
        & 
        \textbf{Ours-RainRobust} & $3.780_{\downarrow0.021}$ & $0.3392_{\uparrow0.0012}$\\
        \midrule
        \multicolumn{1}{c}{\multirow{4}{*}{\tabincell{c}{WeatherStream}}}
         & 
         Restormer & 4.594 & 0.3622\\
        &
        \textbf{Ours-Restormer} & $4.403_{\downarrow0.191}$ & $0.3641_{\uparrow0.0019}$\\
        \cmidrule(l){2-4}
        &
        RainRobust & 3.756 & 0.3309\\
        & 
        \textbf{Ours-RainRobust} & $3.603_{\downarrow0.153}$ & $0.3471_{\uparrow0.0162}$\\
        \bottomrule
    \end{tabularx}
    }
    \centering
\end{table*}

\clearpage
\clearpage

\bibliography{aaai24}


\end{document}

%% file: math_commands.tex
\usepackage{amsmath,amsfonts,bm}

\def\eqref#1{equation~\ref{#1}}

\def\1{\bm{1}}

\DeclareMathAlphabet{\mathsfit}{\encodingdefault}{\sfdefault}{m}{sl}
\SetMathAlphabet{\mathsfit}{bold}{\encodingdefault}{\sfdefault}{bx}{n}

%% file: arxiv.bbl
\begin{thebibliography}{54}
\providecommand{\natexlab}[1]{#1}

\bibitem[{Ancuti et~al.(2019)Ancuti, Ancuti, Sbert, and Timofte}]{ancuti2019dense}
Ancuti, C.~O.; Ancuti, C.; Sbert, M.; and Timofte, R. 2019.
\newblock Dense-haze: A benchmark for image dehazing with dense-haze and haze-free images.
\newblock In \emph{2019 IEEE international conference on image processing (ICIP)}, 1014--1018. IEEE.

\bibitem[{Ancuti et~al.(2021)Ancuti, Ancuti, Vasluianu, and Timofte}]{ancuti2021ntire}
Ancuti, C.~O.; Ancuti, C.; Vasluianu, F.-A.; and Timofte, R. 2021.
\newblock NTIRE 2021 nonhomogeneous dehazing challenge report.
\newblock In \emph{Proceedings of the IEEE/CVF Conference on Computer Vision and Pattern Recognition}, 627--646.

\bibitem[{Ba et~al.(2022)Ba, Zhang, Yang, Suzuki, Pfahnl, Chandrappa, de~Melo, You, Soatto, Wong et~al.}]{ba2022not}
Ba, Y.; Zhang, H.; Yang, E.; Suzuki, A.; Pfahnl, A.; Chandrappa, C.~C.; de~Melo, C.~M.; You, S.; Soatto, S.; Wong, A.; et~al. 2022.
\newblock Not Just Streaks: Towards Ground Truth for Single Image Deraining.
\newblock In \emph{Computer Vision--ECCV 2022: 17th European Conference, Tel Aviv, Israel, October 23--27, 2022, Proceedings, Part VII}, 723--740. Springer.

\bibitem[{Blau et~al.(2018)Blau, Mechrez, Timofte, Michaeli, and Zelnik-Manor}]{blau20182018}
Blau, Y.; Mechrez, R.; Timofte, R.; Michaeli, T.; and Zelnik-Manor, L. 2018.
\newblock The 2018 PIRM challenge on perceptual image super-resolution.
\newblock In \emph{Proceedings of the European Conference on Computer Vision (ECCV) Workshops}, 0--0.

\bibitem[{Chen et~al.(2023{\natexlab{a}})Chen, Ye, Liu, Liao, Jiang, Chen, and Chen}]{10095605}
Chen, S.; Ye, T.; Liu, Y.; Liao, T.; Jiang, J.; Chen, E.; and Chen, P. 2023{\natexlab{a}}.
\newblock MSP-Former: Multi-Scale Projection Transformer for Single Image Desnowing.
\newblock In \emph{ICASSP 2023 - 2023 IEEE International Conference on Acoustics, Speech and Signal Processing (ICASSP)}, 1--5.

\bibitem[{Chen et~al.(2020{\natexlab{a}})Chen, Kornblith, Norouzi, and Hinton}]{chen2020simple}
Chen, T.; Kornblith, S.; Norouzi, M.; and Hinton, G. 2020{\natexlab{a}}.
\newblock A simple framework for contrastive learning of visual representations.
\newblock In \emph{International conference on machine learning}, 1597--1607. PMLR.

\bibitem[{Chen et~al.(2020{\natexlab{b}})Chen, Fang, Ding, Tsai, and Kuo}]{chen2020jstasr}
Chen, W.-T.; Fang, H.-Y.; Ding, J.-J.; Tsai, C.-C.; and Kuo, S.-Y. 2020{\natexlab{b}}.
\newblock JSTASR: Joint size and transparency-aware snow removal algorithm based on modified partial convolution and veiling effect removal.
\newblock In \emph{Computer Vision--ECCV 2020: 16th European Conference, Glasgow, UK, August 23--28, 2020, Proceedings, Part XXI 16}, 754--770. Springer.

\bibitem[{Chen et~al.(2021)Chen, Fang, Hsieh, Tsai, Chen, Ding, Kuo et~al.}]{chen2021all}
Chen, W.-T.; Fang, H.-Y.; Hsieh, C.-L.; Tsai, C.-C.; Chen, I.; Ding, J.-J.; Kuo, S.-Y.; et~al. 2021.
\newblock All snow removed: Single image desnowing algorithm using hierarchical dual-tree complex wavelet representation and contradict channel loss.
\newblock In \emph{Proceedings of the IEEE/CVF International Conference on Computer Vision}, 4196--4205.

\bibitem[{Chen et~al.(2022{\natexlab{a}})Chen, Huang, Tsai, Yang, Ding, and Kuo}]{chen2022learning}
Chen, W.-T.; Huang, Z.-K.; Tsai, C.-C.; Yang, H.-H.; Ding, J.-J.; and Kuo, S.-Y. 2022{\natexlab{a}}.
\newblock Learning multiple adverse weather removal via two-stage knowledge learning and multi-contrastive regularization: Toward a unified model.
\newblock In \emph{Proceedings of the IEEE/CVF Conference on Computer Vision and Pattern Recognition}, 17653--17662.

\bibitem[{Chen et~al.(2023{\natexlab{b}})Chen, Li, Li, and Pan}]{chen2023learning}
Chen, X.; Li, H.; Li, M.; and Pan, J. 2023{\natexlab{b}}.
\newblock Learning A Sparse Transformer Network for Effective Image Deraining.
\newblock In \emph{Proceedings of the IEEE/CVF Conference on Computer Vision and Pattern Recognition}, 5896--5905.

\bibitem[{Chen et~al.(2022{\natexlab{b}})Chen, Pan, Jiang, Li, Huang, Kong, Dai, and Fan}]{chen2022unpaired}
Chen, X.; Pan, J.; Jiang, K.; Li, Y.; Huang, Y.; Kong, C.; Dai, L.; and Fan, Z. 2022{\natexlab{b}}.
\newblock Unpaired deep image deraining using dual contrastive learning.
\newblock In \emph{Proceedings of the IEEE/CVF Conference on Computer Vision and Pattern Recognition}, 2017--2026.

\bibitem[{Chen and Hsu(2013)}]{chen2013generalized}
Chen, Y.-L.; and Hsu, C.-T. 2013.
\newblock A generalized low-rank appearance model for spatio-temporally correlated rain streaks.
\newblock In \emph{Proceedings of the IEEE international conference on computer vision}, 1968--1975.

\bibitem[{Feng et~al.(2023)Feng, Li, Chen, Li, Gu, and Loy}]{feng2023generating}
Feng, R.; Li, C.; Chen, H.; Li, S.; Gu, J.; and Loy, C.~C. 2023.
\newblock Generating Aligned Pseudo-Supervision from Non-Aligned Data for Image Restoration in Under-Display Camera.
\newblock In \emph{Proceedings of the IEEE/CVF Conference on Computer Vision and Pattern Recognition}, 5013--5022.

\bibitem[{Fu et~al.(2017)Fu, Huang, Zeng, Huang, Ding, and Paisley}]{fu2017removing}
Fu, X.; Huang, J.; Zeng, D.; Huang, Y.; Ding, X.; and Paisley, J. 2017.
\newblock Removing rain from single images via a deep detail network.
\newblock In \emph{Proceedings of the IEEE conference on computer vision and pattern recognition}, 3855--3863.

\bibitem[{Garg and Nayar(2007)}]{garg2007vision}
Garg, K.; and Nayar, S.~K. 2007.
\newblock Vision and rain.
\newblock \emph{International Journal of Computer Vision}, 75: 3--27.

\bibitem[{Goodfellow et~al.(2014)Goodfellow, Pouget-Abadie, Mirza, Xu, Warde-Farley, Ozair, Courville, and Bengio}]{goodfellow2014generative}
Goodfellow, I.; Pouget-Abadie, J.; Mirza, M.; Xu, B.; Warde-Farley, D.; Ozair, S.; Courville, A.; and Bengio, Y. 2014.
\newblock Generative adversarial nets.
\newblock \emph{Advances in neural information processing systems}, 27.

\bibitem[{Guo et~al.(2022)Guo, Yan, Anwar, Cong, Ren, and Li}]{guo2022image}
Guo, C.-L.; Yan, Q.; Anwar, S.; Cong, R.; Ren, W.; and Li, C. 2022.
\newblock Image dehazing transformer with transmission-aware 3d position embedding.
\newblock In \emph{Proceedings of the IEEE/CVF Conference on Computer Vision and Pattern Recognition}, 5812--5820.

\bibitem[{Guo et~al.(2021)Guo, Sun, Juefei-Xu, Ma, Xie, Feng, Liu, and Zhao}]{guo2021efficientderain}
Guo, Q.; Sun, J.; Juefei-Xu, F.; Ma, L.; Xie, X.; Feng, W.; Liu, Y.; and Zhao, J. 2021.
\newblock Efficientderain: Learning pixel-wise dilation filtering for high-efficiency single-image deraining.
\newblock In \emph{Proceedings of the AAAI Conference on Artificial Intelligence}, 1487--1495.

\bibitem[{He, Sun, and Tang(2012)}]{he2012guided}
He, K.; Sun, J.; and Tang, X. 2012.
\newblock Guided image filtering.
\newblock \emph{IEEE transactions on pattern analysis and machine intelligence}, 35(6): 1397--1409.

\bibitem[{Hu et~al.(2019)Hu, Fu, Zhu, and Heng}]{hu2019depth}
Hu, X.; Fu, C.-W.; Zhu, L.; and Heng, P.-A. 2019.
\newblock Depth-attentional features for single-image rain removal.
\newblock In \emph{Proceedings of the IEEE/CVF Conference on computer vision and pattern recognition}, 8022--8031.

\bibitem[{Huang, Yu, and He(2021)}]{huang2021memory}
Huang, H.; Yu, A.; and He, R. 2021.
\newblock Memory oriented transfer learning for semi-supervised image deraining.
\newblock In \emph{Proceedings of the IEEE/CVF conference on computer vision and pattern recognition}, 7732--7741.

\bibitem[{Kim et~al.(2013)Kim, Zhang, Andr{\'e}, Chilton, Mackay, Beaudouin-Lafon, Miller, and Dow}]{kim2013cobi}
Kim, J.; Zhang, H.; Andr{\'e}, P.; Chilton, L.~B.; Mackay, W.; Beaudouin-Lafon, M.; Miller, R.~C.; and Dow, S.~P. 2013.
\newblock Cobi: A community-informed conference scheduling tool.
\newblock In \emph{Proceedings of the 26th annual ACM symposium on User interface software and technology}, 173--182.

\bibitem[{Li et~al.(2018)Li, Ren, Fu, Tao, Feng, Zeng, and Wang}]{li2018benchmarking}
Li, B.; Ren, W.; Fu, D.; Tao, D.; Feng, D.; Zeng, W.; and Wang, Z. 2018.
\newblock Benchmarking single-image dehazing and beyond.
\newblock \emph{IEEE Transactions on Image Processing}, 28(1): 492--505.

\bibitem[{Li, Cheong, and Tan(2019)}]{li2019heavy}
Li, R.; Cheong, L.-F.; and Tan, R.~T. 2019.
\newblock Heavy rain image restoration: Integrating physics model and conditional adversarial learning.
\newblock In \emph{Proceedings of the IEEE/CVF conference on computer vision and pattern recognition}, 1633--1642.

\bibitem[{Li, Tan, and Cheong(2020)}]{li2020all}
Li, R.; Tan, R.~T.; and Cheong, L.-F. 2020.
\newblock All in one bad weather removal using architectural search.
\newblock In \emph{Proceedings of the IEEE/CVF conference on computer vision and pattern recognition}, 3175--3185.

\bibitem[{Li et~al.(2022)Li, Zhang, Zhang, Huang, Tian, and Tao}]{li2022toward}
Li, W.; Zhang, Q.; Zhang, J.; Huang, Z.; Tian, X.; and Tao, D. 2022.
\newblock Toward Real-world Single Image Deraining: A New Benchmark and Beyond.
\newblock \emph{arXiv preprint arXiv:2206.05514}.

\bibitem[{Li et~al.(2019)Li, Miao, Ouyang, Ma, Fang, Dong, and Quan}]{li2019lap}
Li, Y.; Miao, Q.; Ouyang, W.; Ma, Z.; Fang, H.; Dong, C.; and Quan, Y. 2019.
\newblock LAP-Net: Level-aware progressive network for image dehazing.
\newblock In \emph{Proceedings of the IEEE/CVF international conference on computer vision}, 3276--3285.

\bibitem[{Li et~al.(2023)Li, Ren, Shu, and Zuo}]{li2023learning}
Li, Y.; Ren, D.; Shu, X.; and Zuo, W. 2023.
\newblock Learning single image defocus deblurring with misaligned training pairs.
\newblock In \emph{Proceedings of the AAAI Conference on Artificial Intelligence}, 1495--1503.

\bibitem[{Liu et~al.(2018)Liu, Jaw, Huang, and Hwang}]{liu2018desnownet}
Liu, Y.-F.; Jaw, D.-W.; Huang, S.-C.; and Hwang, J.-N. 2018.
\newblock DesnowNet: Context-aware deep network for snow removal.
\newblock \emph{IEEE Transactions on Image Processing}, 27(6): 3064--3073.

\bibitem[{Oord, Li, and Vinyals(2018)}]{oord2018representation}
Oord, A. v.~d.; Li, Y.; and Vinyals, O. 2018.
\newblock Representation learning with contrastive predictive coding.
\newblock \emph{arXiv preprint arXiv:1807.03748}.

\bibitem[{Qian et~al.(2018)Qian, Tan, Yang, Su, and Liu}]{qian2018attentive}
Qian, R.; Tan, R.~T.; Yang, W.; Su, J.; and Liu, J. 2018.
\newblock Attentive generative adversarial network for raindrop removal from a single image.
\newblock In \emph{Proceedings of the IEEE conference on computer vision and pattern recognition}, 2482--2491.

\bibitem[{Rai et~al.(2022)Rai, Saluja, Arora, Balasubramanian, Subramanian, and Jawahar}]{rai2022fluid}
Rai, S.~N.; Saluja, R.; Arora, C.; Balasubramanian, V.~N.; Subramanian, A.; and Jawahar, C. 2022.
\newblock Fluid: Few-shot self-supervised image deraining.
\newblock In \emph{Proceedings of the IEEE/CVF Winter Conference on Applications of Computer Vision}, 3077--3086.

\bibitem[{Ronneberger, Fischer, and Brox(2015)}]{ronneberger2015u}
Ronneberger, O.; Fischer, P.; and Brox, T. 2015.
\newblock U-net: Convolutional networks for biomedical image segmentation.
\newblock In \emph{Medical Image Computing and Computer-Assisted Intervention--MICCAI 2015: 18th International Conference, Munich, Germany, October 5-9, 2015, Proceedings, Part III 18}, 234--241. Springer.

\bibitem[{Simonyan and Zisserman(2014)}]{simonyan2014very}
Simonyan, K.; and Zisserman, A. 2014.
\newblock Very deep convolutional networks for large-scale image recognition.
\newblock \emph{arXiv preprint arXiv:1409.1556}.

\bibitem[{Sun et~al.(2018)Sun, Yang, Liu, and Kautz}]{sun2018pwc}
Sun, D.; Yang, X.; Liu, M.-Y.; and Kautz, J. 2018.
\newblock Pwc-net: Cnns for optical flow using pyramid, warping, and cost volume.
\newblock In \emph{Proceedings of the IEEE conference on computer vision and pattern recognition}, 8934--8943.

\bibitem[{Tishby and Zaslavsky(2015)}]{tishby2015deep}
Tishby, N.; and Zaslavsky, N. 2015.
\newblock Deep learning and the information bottleneck principle.
\newblock In \emph{2015 ieee information theory workshop (itw)}, 1--5. IEEE.

\bibitem[{Valanarasu, Yasarla, and Patel(2022)}]{valanarasu2022transweather}
Valanarasu, J. M.~J.; Yasarla, R.; and Patel, V.~M. 2022.
\newblock Transweather: Transformer-based restoration of images degraded by adverse weather conditions.
\newblock In \emph{Proceedings of the IEEE/CVF Conference on Computer Vision and Pattern Recognition}, 2353--2363.

\bibitem[{Wang, Chan, and Loy(2023)}]{wang2023exploring}
Wang, J.; Chan, K.~C.; and Loy, C.~C. 2023.
\newblock Exploring clip for assessing the look and feel of images.
\newblock In \emph{Proceedings of the AAAI Conference on Artificial Intelligence}, 2555--2563.

\bibitem[{Wang et~al.(2023)Wang, Liu, Zhang, Wu, Feng, Zhang, and Zuo}]{wang2023benchmark}
Wang, R.; Liu, X.; Zhang, Z.; Wu, X.; Feng, C.-M.; Zhang, L.; and Zuo, W. 2023.
\newblock Benchmark Dataset and Effective Inter-Frame Alignment for Real-World Video Super-Resolution.
\newblock In \emph{Proceedings of the IEEE/CVF Conference on Computer Vision and Pattern Recognition}, 1168--1177.

\bibitem[{Wang et~al.(2019)Wang, Yang, Xu, Chen, Zhang, and Lau}]{wang2019spatial}
Wang, T.; Yang, X.; Xu, K.; Chen, S.; Zhang, Q.; and Lau, R.~W. 2019.
\newblock Spatial attentive single-image deraining with a high quality real rain dataset.
\newblock In \emph{Proceedings of the IEEE/CVF Conference on Computer Vision and Pattern Recognition}, 12270--12279.

\bibitem[{Wang, Ma, and Liu(2023)}]{wang2023smartassign}
Wang, Y.; Ma, C.; and Liu, J. 2023.
\newblock SmartAssign: Learning a Smart Knowledge Assignment Strategy for Deraining and Desnowing.
\newblock In \emph{Proceedings of the IEEE/CVF Conference on Computer Vision and Pattern Recognition}, 3677--3686.

\bibitem[{Wang, Simoncelli, and Bovik(2003)}]{wang2003multiscale}
Wang, Z.; Simoncelli, E.~P.; and Bovik, A.~C. 2003.
\newblock Multiscale structural similarity for image quality assessment.
\newblock In \emph{The Thrity-Seventh Asilomar Conference on Signals, Systems \& Computers, 2003}, volume~2, 1398--1402. Ieee.

\bibitem[{Wei et~al.(2020)Wei, Liu, Wang, Zhu, Hu, and Zuo}]{wei2020learning}
Wei, Y.; Liu, M.; Wang, H.; Zhu, R.; Hu, G.; and Zuo, W. 2020.
\newblock Learning flow-based feature warping for face frontalization with illumination inconsistent supervision.
\newblock In \emph{Computer Vision--ECCV 2020: 16th European Conference, Glasgow, UK, August 23--28, 2020, Proceedings, Part XII 16}, 558--574. Springer.

\bibitem[{Wei et~al.(2021)Wei, Zhang, Wang, Xu, Yang, Yan, and Wang}]{wei2021deraincyclegan}
Wei, Y.; Zhang, Z.; Wang, Y.; Xu, M.; Yang, Y.; Yan, S.; and Wang, M. 2021.
\newblock Deraincyclegan: Rain attentive cyclegan for single image deraining and rainmaking.
\newblock \emph{IEEE Transactions on Image Processing}, 30: 4788--4801.

\bibitem[{Wu et~al.(2021)Wu, Qu, Lin, Zhou, Qiao, Zhang, Xie, and Ma}]{wu2021contrastive}
Wu, H.; Qu, Y.; Lin, S.; Zhou, J.; Qiao, R.; Zhang, Z.; Xie, Y.; and Ma, L. 2021.
\newblock Contrastive learning for compact single image dehazing.
\newblock In \emph{Proceedings of the IEEE/CVF Conference on Computer Vision and Pattern Recognition}, 10551--10560.

\bibitem[{Xia et~al.(2023)Xia, Monica, Chao, Hariharan, Weinberger, and Campbell}]{xia2023image}
Xia, Y.; Monica, J.; Chao, W.-L.; Hariharan, B.; Weinberger, K.~Q.; and Campbell, M. 2023.
\newblock Image-to-Image Translation for Autonomous Driving from Coarsely-Aligned Image Pairs.
\newblock In \emph{2023 IEEE International Conference on Robotics and Automation (ICRA)}, 7756--7762. IEEE.

\bibitem[{Yasarla, Sindagi, and Patel(2021)}]{yasarla2021semi}
Yasarla, R.; Sindagi, V.~A.; and Patel, V.~M. 2021.
\newblock Semi-supervised image deraining using gaussian processes.
\newblock \emph{IEEE Transactions on Image Processing}, 30: 6570--6582.

\bibitem[{Ye et~al.(2022)Ye, Yu, Chang, Zhu, Zhao, Yan, and Tian}]{ye2022unsupervised}
Ye, Y.; Yu, C.; Chang, Y.; Zhu, L.; Zhao, X.-l.; Yan, L.; and Tian, Y. 2022.
\newblock Unsupervised Deraining: Where Contrastive Learning Meets Self-similarity.
\newblock In \emph{Proceedings of the IEEE/CVF Conference on Computer Vision and Pattern Recognition}, 5821--5830.

\bibitem[{Zamir et~al.(2022)Zamir, Arora, Khan, Hayat, Khan, and Yang}]{zamir2022restormer}
Zamir, S.~W.; Arora, A.; Khan, S.; Hayat, M.; Khan, F.~S.; and Yang, M.-H. 2022.
\newblock Restormer: Efficient transformer for high-resolution image restoration.
\newblock In \emph{Proceedings of the IEEE/CVF conference on computer vision and pattern recognition}, 5728--5739.

\bibitem[{Zhang et~al.(2023)Zhang, Ba, Yang, Mehra, Gella, Suzuki, Pfahnl, Chandrappa, Wong, and Kadambi}]{zhang2023weatherstream}
Zhang, H.; Ba, Y.; Yang, E.; Mehra, V.; Gella, B.; Suzuki, A.; Pfahnl, A.; Chandrappa, C.~C.; Wong, A.; and Kadambi, A. 2023.
\newblock Weatherstream: Light transport automation of single image deweathering.
\newblock In \emph{Proceedings of the IEEE/CVF Conference on Computer Vision and Pattern Recognition}, 13499--13509.

\bibitem[{Zhang and Patel(2018)}]{zhang2018density}
Zhang, H.; and Patel, V.~M. 2018.
\newblock Density-aware single image de-raining using a multi-stream dense network.
\newblock In \emph{Proceedings of the IEEE conference on computer vision and pattern recognition}, 695--704.

\bibitem[{Zhang et~al.(2021{\natexlab{a}})Zhang, Li, Yu, Luo, and Li}]{zhang2021deep}
Zhang, K.; Li, R.; Yu, Y.; Luo, W.; and Li, C. 2021{\natexlab{a}}.
\newblock Deep dense multi-scale network for snow removal using semantic and depth priors.
\newblock \emph{IEEE Transactions on Image Processing}, 30: 7419--7431.

\bibitem[{Zhang et~al.(2021{\natexlab{b}})Zhang, Wang, Liu, Wang, Zhang, and Zuo}]{zhang2021learning}
Zhang, Z.; Wang, H.; Liu, M.; Wang, R.; Zhang, J.; and Zuo, W. 2021{\natexlab{b}}.
\newblock Learning raw-to-srgb mappings with inaccurately aligned supervision.
\newblock In \emph{Proceedings of the IEEE/CVF International Conference on Computer Vision}, 4348--4358.

\bibitem[{Zhang et~al.(2022)Zhang, Wang, Zhang, Chen, and Zuo}]{zhang2022self}
Zhang, Z.; Wang, R.; Zhang, H.; Chen, Y.; and Zuo, W. 2022.
\newblock Self-supervised learning for real-world super-resolution from dual zoomed observations.
\newblock In \emph{European Conference on Computer Vision}, 610--627. Springer.

\end{thebibliography}
